\documentclass[11pt]{article}
\usepackage[margin=1in]{geometry}
\usepackage[T1]{fontenc}
\usepackage{lmodern}
\usepackage{microtype}
\usepackage{amsmath,amssymb,amsthm}
\usepackage{booktabs}
\usepackage{graphicx}
\usepackage{xcolor}
\usepackage{algorithm}
\usepackage{algpseudocode}
\usepackage[numbers,sort&compress]{natbib}
\usepackage[colorlinks=true,linkcolor=blue!50!black,citecolor=blue!50!black,urlcolor=blue!50!black]{hyperref}
\usepackage[capitalise,noabbrev]{cleveref}
\usepackage{enumitem}
\usepackage{float}
\setlist{itemsep=2pt,topsep=3pt}
\definecolor{newpart}{RGB}{0,120,120}
\newcommand{\newpart}[1]{\textcolor{newpart}{#1}}

\newtheorem{proposition}{Proposition}
\theoremstyle{remark}

\newcommand{\R}{\mathbb{R}}
\newcommand{\taunear}{\tau_{\mathrm{near}}}
\newcommand{\taucount}{\tau_{\mathrm{count}}}
\newcommand{\mhat}{\widehat{\mathrm{MSE}}}
\newcommand{\nearest}{\operatorname{nearest}}
\newcommand{\tlast}[1]{t^{\mathrm{last}}_{#1}}
\newcommand{\nfar}{n_{\mathrm{far}}}
\newcommand{\Count}{\textsc{count}}
\newcommand{\Off}{\textsc{off}}
\newcommand{\Dist}{\textsc{distance}}
\newcommand{\Est}{\textsc{estimate}}
\DeclareMathOperator*{\argmin}{arg\,min}

\newcommand{\NeverReachCount}{9}

\newcommand{\NearReachCount}{7}
\newcommand{\NearReachWithinTwo}{5}

\newcommand{\CovOffBelowLloyd}{24}

\newcommand{\CovCountAboveLloyd}{58}
\newcommand{\CovCountAboveLloydTen}{17}

\newcommand{\ExactBetterMsdWhere}{k-means++ initialization, $k=1024$, 5 epochs}

\newcommand{\MsdWorstWhere}{k-means++ initialization, $k=4096$, 10 epochs}

\newcommand{\AbandonRuns}{800}
\newcommand{\AbandonRunsHit}{75}
\newcommand{\AbandonRemovals}{253}
\newcommand{\AbandonAllRemovals}{136{,}890}
\newcommand{\FracAboveFourKdd}{23--24}
\newcommand{\FracAboveFourCensus}{13--23}
\newcommand{\FracAboveFourMsd}{33--43}
\newcommand{\FracAboveFourGeo}{69--73}
\newcommand{\FracAboveFourCov}{38--45}

\newcommand{\TimeDeepLib}{32.9}
\newcommand{\TimeDeepNear}{32.9}
\newcommand{\TimeDeepExact}{90.3}
\newcommand{\TimeKddLib}{11.9}
\newcommand{\TimeKddNear}{12.0}
\newcommand{\TimeKddExact}{27.6}
\newcommand{\NearRatioMin}{0.98}
\newcommand{\NearRatioMax}{1.04}
\newcommand{\TimingReps}{6}

\newcommand{\EmbRuns}{103}
\newcommand{\HolmFamilySizeHeld}{272}
\newcommand{\HolmFamilySizeDev}{120}

\title{Small Is Not Dead: Distance-Aware Cluster Revival for e-means}
\author{Eyal Weiss\\ Robotics Institute, Carnegie Mellon University}
\date{}

\begin{document}
\maketitle

\begin{abstract}
e-means~\citep{bessa2026emeans} is a recent mini-batch k-means algorithm that reaches the error of Lloyd's algorithm about an order of magnitude faster on large embedding datasets, and it is now part of the FAISS library. In addition to its moving-average update of the centres, e-means periodically relocates the centres of small clusters: every five mini-batches, every cluster whose running point count falls below a tenth of the average count is removed, and its centre is reused to split a large cluster. In the authors' ablation, removing this mechanism, called dead-cluster revival, raises the error more than removing any other component added to the moving-average update, and without it many centres stop receiving points. The rule, however, considers only cluster \textsl{size}. It cannot distinguish a cluster that is small because it is redundant from one that is small because it covers a rare, isolated group of points, and on data with such groups it repeatedly removes centres whose removal is costly. We add a second condition, based on a second source of information, the distance between centres: a small cluster is removed only if its centre is close to another centre, as measured by the squared distance between the two centres relative to a running estimate of the mean squared error. When a centre equals the mean of its points, this squared distance is an upper bound on the average increase in squared error, per point of the cluster, caused by removing the centre. The condition costs one nearest-centre search for the small clusters every five mini-batches, and in our timings it adds no measurable running time. On the embedding benchmarks of the e-means paper, it never prevents a removal, so the results are identical to those of the FAISS implementation. On datasets with small isolated groups that were not used during development, it reduces the test error after two epochs (passes over the data) by roughly 10\% to 80\% relative to count-based revival, depending on the dataset and the number of centres, and it preserves the advantage of k-means++ initialization, which count-based revival loses on these datasets. On the other evaluation datasets, its test error after two epochs is within about 1\% of the better of revival enabled and revival disabled. We also report two failure modes that the analysis predicts: the condition ignores the error reduction obtained by the split that replaces a removed cluster, and it compares distances with a single, global error scale. This work is a CPU proof of concept; GPU and multi-GPU implementations are left to future work.
\end{abstract}

\section{Introduction}

The k-means problem asks for $k$ centres that minimize the mean squared distance from the points of a dataset to their nearest centre. It is a standard tool for vector quantization, data summarization and approximate nearest-neighbour search~\citep{douze2024faiss}. Lloyd's algorithm~\citep{lloyd1982}, which alternates between assigning every point to its nearest centre and moving every centre to the mean of the points assigned to it, remains the default method for solving it.

e-means~\citep{bessa2026emeans} is a recent replacement for Lloyd's algorithm that reaches the same error about an order of magnitude faster. It was designed for, and evaluated on, large collections of embedding vectors: on these data it reaches the test error of 25 Lloyd iterations in fewer than three epochs, where an epoch is one complete pass over the training data, and the authors report a reduction in GPU running time of an order of magnitude or more compared with state-of-the-art k-means implementations (1.4--2.6 epochs and a 10--45-fold reduction~\citep[Table~4]{bessa2026emeans}). e-means processes the data in mini-batches and represents each centre as the ratio of two exponential moving averages (EMAs), of the number and of the sum of the points assigned to it; a decreasing step-size schedule and two scaling laws, empirical formulas that set the batch size and the step size from the size of the problem, remove the need for tuning (\cref{sec:background}). Since version 1.16, e-means is distributed with the FAISS library~\citep{douze2024faiss}; we refer to its implementation there as \textsl{the library}.

In addition to the moving-average update, e-means relocates the centres of small clusters during training, a mechanism that the authors call dead-cluster revival. Every five mini-batches, every cluster whose moving-average count has fallen below a tenth of the average count is removed, and its centre is reused to split a large cluster into two. Without this mechanism, many centres stop receiving points: on the DEEP dataset ($10^6$ points, $k=16{,}384$), about a quarter of the centres end with fewer than ten points when revival is disabled, against almost none with revival (26\% and 0.3\%~\citep[Table~14]{bessa2026emeans}). The authors' ablation removes one component at a time while keeping the moving-average update, which is the core of the method. Among the components it removes, revival is the one whose removal costs most: on the same data, removing it raises the train and test errors by a few percent after three epochs, whereas removing the step-size schedule or the final averaging step changes them by only a fraction of a percent (3.0\% and 2.2\%, against at most 0.26\%~\citep[Table~1]{bessa2026emeans}). On embedding data, where the clusters that become small are mostly redundant, moving their centres into large clusters reduces the error and makes the cluster sizes more uniform.

The revival rule decides by cluster size alone, but \textsl{the size of a cluster does not determine the cost of removing its centre}. A cluster can be small because it is redundant, or because it covers a small group of points far from all other points: a rare category in a tabular dataset, an unusual operating regime of a sensor, or a remote town in a geographic dataset. Removing the centre of a redundant cluster increases the error very little. Removing the centre of an isolated group moves its points to a distant centre, and each of these points then contributes its squared distance to that centre. We found that on such data e-means repeatedly removes exactly these centres. On the UCI Covertype dataset with standardized features, e-means with $k=256$ ends \CovCountAboveLloyd\% above the test error of 25 Lloyd iterations after two epochs, and still \CovCountAboveLloydTen\% above it after ten epochs, whereas with revival disabled it ends \CovOffBelowLloyd\% below Lloyd's error after two epochs (\cref{sec:failure}).

The authors of e-means have also encountered this problem. A recent change to the library disables revival by default for data with at most two dimensions, after the authors found that disabling it lowered the median test error on two-dimensional geographic datasets (by 7--10\%~\citep{faiss2026pr5854}). This change is documented in the library's source code and is not described in the e-means paper. It is a coarse mitigation: it decides by dimension, whereas the failure depends on whether small clusters are isolated, which also occurs in high dimension, for example on Covertype (54 dimensions) and KDD Cup 99 (122 dimensions after encoding).

\subsection{Our Contribution}

We propose a better-informed revival rule. It keeps the size test of e-means and adds a second condition that uses a second source of information, the distance from the centre of a small cluster to the nearest other centre: \textsl{a small cluster is removed only if its centre is close to another centre}. Clusters that are small because they are redundant are therefore still removed, and clusters that cover isolated groups of points are kept. In our timings, the added condition has no measurable computational cost.

\paragraph{Rule.} A cluster $j$ selected by the size test, which we call a \textsl{candidate}, is removed only if $\delta_j=\min_{i\ne j}\|c_j-c_i\|^2$, the squared distance from its centre $c_j$ to the nearest other centre, is at most $\taunear\,\mhat$, where $\mhat$ is a running estimate of the mean squared error (MSE) and $\taunear$ is a threshold; or if it has received no point for two epochs (\cref{sec:method}). The distance test has a simple interpretation. When $c_j$ is the mean of the points nearest to it, $\delta_j$ is an upper bound on the average increase in squared error, per point of the cluster, caused by removing $c_j$ while the other centres are kept fixed (\cref{prop:bound}); summed over the points of the cluster, this increase is Fritzke's utility~\citep{fritzke1997lbgu,fritzke2017kmeansu}. Under these conditions, the test removes a candidate only if removing it increases the squared error of its points, on average, by at most $\taunear$ times the current MSE. We use $\taunear=4$, a margin above the largest values of $\delta_j/\mhat$ that we observed among the candidates on embedding data, where removals are beneficial. The exact value matters little: on held-out data, values from 2 to 8 give nearly the same results (\cref{sec:exp-sens}), so the threshold only needs to lie roughly in this range. The second condition, which we call the \textsl{abandonment clause}, releases a centre that has lost all its points while being far from all other centres. It is a safeguard motivated by an argument rather than by an observed failure: such a centre can be removed at no cost, but the distance test alone would keep it until the end of training. In our experiments, it rarely acts, and removing it changes the results negligibly (\cref{sec:exp-sens}); the gains come from the distance test. Two epochs is the shortest waiting period, measured in whole epochs, after which this clause never removes a centre that receives at least one point in every epoch, whatever the batch size: when the data are shuffled before every epoch, a point visited at the start of one epoch may next be visited at the end of the following epoch, almost two epochs later (\cref{prop:window}).

\paragraph{Cost.} Count-based revival already keeps a moving-average count per cluster and checks it every five batches. Our rule adds one moving average per cluster, computed from the point-to-centre distances that the assignment step already returns, and a nearest-centre search for the candidates at each check. In the two configurations that we timed, the running time with our rule was the same as that of the library, within the variation between repeated runs (\cref{sec:exp-cost}).

\paragraph{Evidence.} We follow an experimental protocol that was written down before the runs, and we evaluate on held-out datasets that were not used during development (\cref{sec:experiments}). On three embedding datasets of the e-means paper, in four of its configurations with up to ten million points and $65{,}536$ centres, the distance test never prevents a removal, so all runs are identical to those of the library. On held-out datasets with small isolated groups, it reduces the test error after two epochs relative to count-based revival by roughly 25--80\% on KDD Cup 99, 20--75\% on Shuttle and 10--25\% on GeoNames, depending on the number of centres. On the other evaluation datasets, its test error after two epochs is within about 1\% of the better of revival enabled and revival disabled. With k-means++ initialization, it keeps the centres that k-means++ places on isolated groups, which count-based revival removes: on KDD Cup 99 ($k=1024$, two epochs), count-based revival ends at more than 25 times the test error of 25 Lloyd iterations started from the same centres, and our rule within 1\% of it. Compared with an online estimate of the utility, the bound is as good or better in most configurations, and much better in training error on data with rare points far from all others; the estimate is better on YearPredictionMSD. The bound adds no measurable running time and uses only the centres, whereas the estimate requires a search for the two nearest centres of every point in every batch. Finally, we document two failure modes that the analysis predicts (\cref{sec:exp-mech}).

We release the code.\footnote{\url{https://github.com/eyal-weiss/emeans-distance-revival}} It is a subclass of the library's NumPy implementation and can replace it without other changes.

\section{Background: e-means and Count-Based Revival}\label{sec:background}

Let $X=\{x_1,\dots,x_n\}\subset\R^d$ be the training points and $c_1,\dots,c_k\in\R^d$ the centres. We write $\nearest(x)=\argmin_j\|x-c_j\|^2$ for the index of the centre nearest to a point $x$, and say that $x$ is assigned to that centre. The k-means objective is the mean squared error (MSE), $\frac1n\sum_{i=1}^n\min_j\|x_i-c_j\|^2$; the sum of squared errors (SSE) is $n$ times the MSE. Lloyd's algorithm~\citep{lloyd1982} alternates between assigning every point to its nearest centre and replacing every centre by the mean of the points assigned to it; each such iteration is one epoch. As in the e-means paper, our reference is FAISS's implementation of Lloyd's algorithm run for 25 iterations, which we denote Lloyd@25.

\paragraph{The e-means update.} e-means~\citep[Sec.~3]{bessa2026emeans} processes the data in epochs. Each epoch visits every training point once, in a new random order, in batches of $B$ points, so that an epoch consists of $L=\lceil n/B\rceil$ batches. We denote by $T$ the number of epochs, by $M=TL$ the total number of batches, and by $t\in\{0,\dots,M-1\}$ the index of the current batch. For each cluster $j$, e-means keeps two EMAs, a count $N_j\in\R_+$ and a sum $S_j\in\R^d$ of the points assigned to $j$, and defines the centre as their ratio. With $n_j$ and $s_j$ the number and the sum of the points $x$ of batch $t$ with $\nearest(x)=j$,
\begin{equation}\label{eq:ema}
N_j \leftarrow (1-\alpha_t)N_j + \alpha_t n_j,\qquad S_j \leftarrow (1-\alpha_t)S_j + \alpha_t s_j,\qquad c_j=S_j/N_j,
\end{equation}
where $\alpha_t\in(0,1]$ is the step size (the learning rate). Equivalently, $c_j$ moves towards the mean $s_j/n_j$ of its batch points by a fraction $\alpha_t n_j/N_j\in[0,1]$ of the distance, where $N_j$ is the updated count; a cluster that receives no point does not move. With the whole dataset as one batch and $\alpha_t=1$, \cref{eq:ema} is one iteration of Lloyd's algorithm. The step size decreases from $\alpha_0$ to $0.1\,\alpha_0$ along a cosine curve, $\alpha_t=\alpha_0\big(0.1+0.45\,(1+\cos(\pi t/M))\big)$, a schedule known as cosine annealing~\citep{loshchilov2017sgdr}. The batch size $B$ and the initial step size $\alpha_0$ are set by two scaling laws, that is, empirical power laws in $n$, $k$ and $T$, which the authors fitted on one embedding dataset~\citep[Sec.~3.3]{bessa2026emeans}; the library's default (\texttt{batch\_law="auto"}) uses a smaller batch size for $d\le16$. The centres are initialized at $k$ training points drawn at random, with a negligible initial count. After the last epoch, each centre is replaced by the mean of the points assigned to it during that epoch, a step that the authors call the \textsl{last-epoch mean}. The EMA update itself comes from vector-quantized autoencoders~\citep{vandenoord2017vqvae}; e-means combines it with the step-size schedule, the scaling laws, the last-epoch mean and the revival rule described below.

\paragraph{Count-based revival.} Every $R=5$ batches, the clusters whose count is below a fraction $\taucount=0.1$ of the average count, $N_j<\taucount\overline{N}$ with $\overline{N}=\frac1k\sum_{i=1}^k N_i$, form the candidate set $\mathcal{C}$. Each candidate is removed, and its centre is reused to split a \textsl{donor} cluster. Donors are drawn without replacement among the non-candidates, with probability proportional to their count. The donor's centre $c$ is replaced by $c\odot(1-\epsilon\sigma)$, and the candidate's centre is set to $c\odot(1+\epsilon\sigma)$, where $\odot$ denotes the elementwise product, $\sigma=(1,-1,1,-1,\dots)$ and $\epsilon=10^{-4}$; both clusters receive half of the donor's count, with sums that match their new centres. FAISS's implementation of Lloyd's algorithm applies the same split only to clusters that have become empty, whereas e-means also removes clusters that still receive points, at any time during training. We call this rule \textsl{count-based revival}.

\section{When Count-Based Revival Fails}\label{sec:failure}

Consider a cluster $j$ whose centre covers a group of $\nfar$ points that lie at distance about $\rho$ from every other centre, with $\rho$ large compared with the spread of the group. The group contributes about $\nfar B/n$ points to a batch, while the average count $\overline{N}$ is about $B/k$; so if $\nfar<\taucount\, n/k$, cluster $j$ is typically a candidate at every check. Removing it moves all $\nfar$ points to a centre at squared distance about $\rho^2$, which increases the SSE by about $\nfar\rho^2$. This increase can exceed the SSE of many ordinary clusters combined, while the split that reuses the centre reduces the error much less. Two properties of e-means make this worse. First, $N_j$ is a moving average over a few batches, so a group whose points appear in only a few batches of each epoch (for example, a handful of rare points) has $N_j$ close to zero between these batches. Second, the rule is applied until the end of training, so a centre that moves back to the group is removed again.

\begin{figure}[t]
\centering
\includegraphics[width=0.8\linewidth]{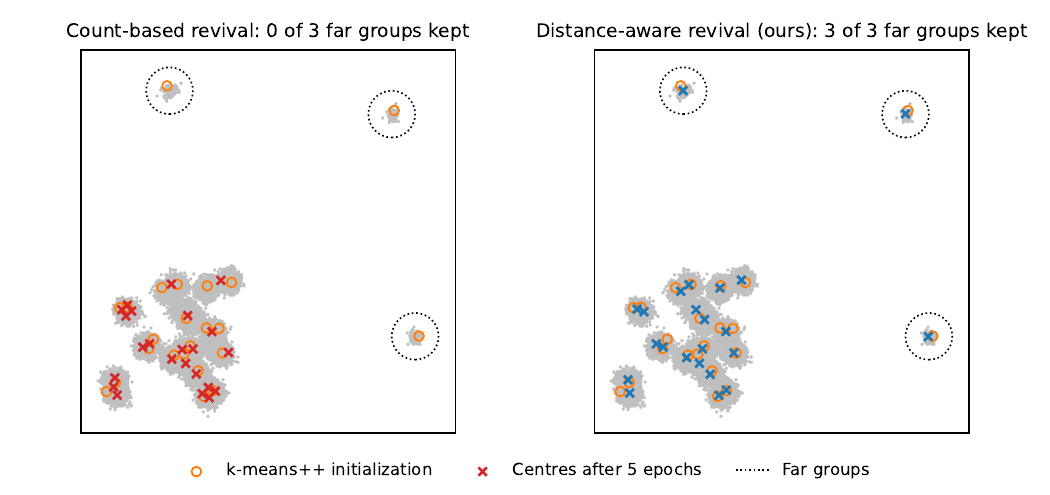}
\caption{Two-dimensional example: 20 Gaussian blobs of 2{,}000 points and three far groups of 80 points (dotted circles), $k=23$ centres initialized by k-means++ (orange circles), and 5 epochs of e-means. Left: count-based revival removes the centres of the three far groups and reuses them inside the main mass of points. Right: with the distance test, all three are kept. Seed 0 is shown; the outcome (0 of 3 and 3 of 3 far groups kept) is the same in all 10 seeds. The size threshold is the e-means paper's $\taucount=0.1$; since version 1.16, the library disables revival by default for $d\le2$.}
\label{fig:failure}
\end{figure}

\cref{fig:failure} shows the effect in two dimensions. Twenty Gaussian blobs of 2{,}000 points each are complemented by three far groups of 80 points each (4\% of the size of a typical cluster), and $k=23$. k-means++ initialization~\citep{arthur2007kmeanspp} places a centre on every far group in all 10 random seeds that we ran. After 5 epochs, count-based revival has removed all three of these centres in all 10 seeds, whereas with the distance test of \cref{sec:method} all three are kept in all 10 seeds.

The same failure occurs on real data. On UCI Covertype with standardized features (each feature shifted and scaled to zero mean and unit variance), each of the 40 soil types is encoded as a binary (one-hot) feature. For the rarest soil types, standardization maps the value 1 to hundreds of standard deviations from the mean, so the few points with such a soil type form small, isolated groups. With $k=256$, e-means with its default settings ends \CovCountAboveLloyd\% above the test error of Lloyd@25 after 2 epochs, and disabling revival brings it to \CovOffBelowLloyd\% below Lloyd@25 (\cref{sec:experiments}). \textsl{The gap is therefore caused by the removal rule, not by the moving-average optimization.}

\section{Distance-Aware Revival}\label{sec:method}

We keep the candidate set $\mathcal{C}$ of count-based revival and add a condition that decides which candidates are removed: a candidate is removed only if its centre is close to another centre, or if it has received no point for a long time. Each cluster $j$ carries one more EMA, $D_j$, of its batch distortion, that is, of the sum of the squared distances from the batch points assigned to $j$ to $c_j$; these distances are available from the assignment step, and $D_j$ is updated with the same step size as $N_j$. Each cluster also records $\tlast{j}$, the index of the last batch in which it received a point, or of the batch in which a split created it; the latter gives a newly created centre a full waiting period before it can be considered abandoned. At a revival check at batch $t$,
\begin{equation}\label{eq:rule}
\mhat=\frac{\sum_i D_i}{\sum_i N_i},\qquad \delta_j=\min_{i\ne j}\|c_j-c_i\|^2,\qquad
\text{remove } j\in\mathcal{C} \iff \delta_j \le \taunear\, \mhat \ \text{ or } \ t-\tlast{j} \ge W L,
\end{equation}
with $L=\lceil n/B\rceil$ batches per epoch. Here $\mhat$, the ratio of the two moving averages, is a running estimate of the MSE over all clusters, $\delta_j$ is the squared distance from $c_j$ to the nearest other centre, $\taunear$ is a threshold, and $W$ is a waiting period measured in epochs. We call the second condition the abandonment clause. It handles a case that the distance test cannot: removing a centre that has lost all its points does not change the SSE, but if that centre is far from every other centre, the distance test keeps it until the end of training. The clause is motivated by this argument, not by a failure that we observed; in our experiments, it rarely acts and has no measurable effect on the error (\cref{sec:exp-sens}). Removed candidates are reused exactly as in the library, with the same random procedure for choosing donors; the other candidates keep their centres and statistics. We use $\taunear=4$ and $W=2$ throughout; both values were chosen on the development data and fixed before the evaluation on the held-out datasets (\cref{sec:experiments}). \cref{alg:rule} gives the pseudo-code, with the parts that differ from the library in \newpart{colour}.

\begin{algorithm}[t]
\caption{Revival in e-means with the distance test. The parts in \newpart{colour} are added by our rule; the rest is the library's procedure, unchanged.}\label{alg:rule}
\begin{algorithmic}[1]
\Statex \textbf{Initialization}
\State \newpart{$D_j\gets0$ and $\tlast{j}\gets0$ for every cluster $j$}
\Statex \textbf{After each batch $t$}
\State update $N_j$ and $S_j$ as in e-means (\cref{eq:ema})
\State \newpart{$D_j \gets (1-\alpha_t)D_j+\alpha_t\sum_{x\in\text{batch}:\, \nearest(x)=j}\|x-c_j\|^2$}
\State \newpart{\textbf{if} $n_j>0$ \textbf{then} $\tlast{j}\gets t$}
\Statex \textbf{Every $R=5$ batches}
\State $\mathcal{C}\gets\{j: N_j<\taucount\overline{N}\}$ \Comment{size test}
\State \newpart{$\mhat\gets\sum_i D_i/\sum_i N_i$}
\State \newpart{$\delta_j\gets\min_{i\ne j}\|c_j-c_i\|^2$ for each $j\in\mathcal{C}$} \Comment{nearest-centre search}
\State \newpart{$\mathcal{C}_{\mathrm{remove}}\gets\{j\in\mathcal{C}: \delta_j\le\taunear \mhat \ \text{or}\ t-\tlast{j}\ge WL\}$} \Comment{library: $\mathcal{C}_{\mathrm{remove}}=\mathcal{C}$}
\For{each $j\in\mathcal{C}_{\mathrm{remove}}$}
\State draw a new donor among the non-candidates, with probability proportional to $N$
\State split the donor's cluster, and set $c_j$ to one of the two halves
\State \newpart{split $D$ like $N$, and set $\tlast{j}\gets t$}
\EndFor
\end{algorithmic}
\end{algorithm}

\begin{proposition}[Removal cost]\label{prop:bound}
Let $P_j$ be the set of training points whose nearest centre is $c_j$, assumed non-empty, let $\mu_j$ be their mean, and let $c'$ be a centre nearest to $c_j$ among the others, so that $\delta_j=\|c_j-c'\|^2$. With the other centres kept fixed, removing $c_j$ increases the sum of squared errors by
\[
U_j=\sum_{x\in P_j}\Big(\min_{i\ne j}\|x-c_i\|^2-\|x-c_j\|^2\Big)\ \le\ |P_j|\big(\delta_j+2\langle\mu_j-c_j,\,c_j-c'\rangle\big)\ \le\ |P_j|\big(\delta_j+2\|\mu_j-c_j\|\sqrt{\delta_j}\big),
\]
with equality in the first inequality if and only if $c'$ is a nearest remaining centre of every point of $P_j$. In particular, if $c_j=\mu_j$ (as at a fixed point of Lloyd's algorithm), then $U_j\le|P_j|\,\delta_j$.
\end{proposition}
\begin{proof}
We first replace the nearest remaining centre of each point by $c'$, then compute the resulting increase exactly, point by point, and sum over $P_j$.

\emph{Step 1 (one replacement centre for all points).} Let $x\in P_j$. Since $c'$ is one of the remaining centres, $\min_{i\ne j}\|x-c_i\|^2\le\|x-c'\|^2$, with equality if and only if $c'$ is a nearest remaining centre of $x$. Subtracting $\|x-c_j\|^2$ from both sides and summing over $P_j$ gives
\begin{equation}\label{eq:prop1-replace}
U_j\le\sum_{x\in P_j}\big(\|x-c'\|^2-\|x-c_j\|^2\big),
\end{equation}
with equality if and only if $c'$ is a nearest remaining centre of every point of $P_j$.

\emph{Step 2 (one point).} Write $x-c'=(x-c_j)+(c_j-c')$ and expand the square:
\[
\|x-c'\|^2=\|x-c_j\|^2+2\langle x-c_j,\,c_j-c'\rangle+\|c_j-c'\|^2 .
\]
Since $\|c_j-c'\|^2=\delta_j$, the increase for the point $x$ is
\begin{equation}\label{eq:prop1-point}
\|x-c'\|^2-\|x-c_j\|^2=\delta_j+2\langle x-c_j,\,c_j-c'\rangle .
\end{equation}

\emph{Step 3 (sum over the cluster).} Summing \cref{eq:prop1-point} over the $|P_j|$ points of $P_j$, and using the linearity of the inner product,
\[
\sum_{x\in P_j}\big(\|x-c'\|^2-\|x-c_j\|^2\big)=|P_j|\,\delta_j+2\Big\langle\sum_{x\in P_j}(x-c_j),\,c_j-c'\Big\rangle .
\]
Since $\mu_j$ is the mean of $P_j$, $\sum_{x\in P_j}x=|P_j|\,\mu_j$, so $\sum_{x\in P_j}(x-c_j)=|P_j|(\mu_j-c_j)$. Therefore
\begin{equation}\label{eq:prop1-sum}
\sum_{x\in P_j}\big(\|x-c'\|^2-\|x-c_j\|^2\big)=|P_j|\big(\delta_j+2\langle\mu_j-c_j,\,c_j-c'\rangle\big).
\end{equation}
Combining \cref{eq:prop1-replace,eq:prop1-sum} gives the first inequality, with its equality condition.

\emph{Step 4 (Cauchy--Schwarz).} By the Cauchy--Schwarz inequality, $\langle\mu_j-c_j,\,c_j-c'\rangle\le\|\mu_j-c_j\|\,\|c_j-c'\|=\|\mu_j-c_j\|\sqrt{\delta_j}$, which gives the second inequality.

\emph{Step 5 (centre at the mean).} If $c_j=\mu_j$, the inner product in \cref{eq:prop1-sum} is zero, and the first inequality becomes $U_j\le|P_j|\,\delta_j$.
\end{proof}
$U_j$ is Fritzke's utility of centre $j$~\citep{fritzke1997lbgu,fritzke2017kmeansu}: the increase in SSE caused by removing it while the other centres are kept fixed. When $c_j=\mu_j$, $|P_j|\delta_j$ also upper-bounds the increase in SSE caused by merging $P_j$ with the cluster $P'$ of $c'$, $\tfrac{|P_j||P'|}{|P_j|+|P'|}\delta_j$, which is the merge cost of Ward's agglomerative clustering and of the pairwise nearest neighbour (PNN) method~\citep{ward1963,equitz1989}. For a candidate ($|P_j|\ll|P'|$) the two quantities are nearly equal. \cref{prop:bound} motivates the test; it does not make the rule exact. Under its conditions, \textsl{removing a candidate increases the squared error of its points, on average, by at most $\taunear$ times the current MSE}; dividing by $\mhat$ makes the test invariant to the scale of the data. Individual points can lose more, up to $\delta_j+2r\sqrt{\delta_j}$ for a cluster of radius $r$ (the largest distance from $c_j$ to a point of $P_j$). The conditions are that one centre is removed while the others are kept fixed, and that the centre equals the mean of its points. Three aspects of the algorithm violate them. First, two nearby candidates can each pass the test because of the other, and both are then removed; breathing k-means avoids the analogous problem by excluding the nearest neighbour of a centre selected for removal from removal in the same step, which its author calls freezing~\citep{fritzke2020breathing}, and we do not. Second, a centre can lag behind its points, as described next. Third, the abandonment clause removes a candidate regardless of distance.

\paragraph{Centre lag.} In e-means, $c_j=S_j/N_j$ is a weighted mean, with exponentially decaying weights, of the points assigned to $j$ in past batches, together with its initial or split position; it is not the mean of the points currently nearest to it. For that weighted set of points, $c_j$ is the mean, so the weighted version of the inner product in \cref{eq:prop1-sum} is zero, as in Step~5 of the proof. The points currently nearest to $c_j$ can have a mean $\mu_j\neq c_j$, and \cref{prop:bound} then adds $2\langle\mu_j-c_j,c_j-c'\rangle$ per point. This term is positive when $c_j$ lags behind a group of points that lies on the side away from $c'$; $\delta_j$ then underestimates the cost, and the test errs towards removal, which is the library's behaviour.

\paragraph{Bound versus utility estimate.} The bound uses only the centres; it needs no distance from each point to its second-nearest centre. We compare it in \cref{sec:exp-exact} with an online estimate of the utility per point: an EMA of $\sum(d_2^2-d_1^2)$ over the batch points assigned to $j$, divided by the matching EMA of their number, where $d_1$ is the distance from a point to its assigned centre and $d_2$ the distance to its nearest other centre. Because it averages over past batches, in which the centres were at different positions, it estimates $U_j/|P_j|$ rather than equals it, and it requires a search for the two nearest centres of every point in every batch.

\begin{proposition}[Waiting period]\label{prop:window}
Suppose that the run consists of whole epochs, each visiting every training point exactly once in $L=\lceil n/B\rceil$ batches (as e-means does when the number of epochs is an integer, as in all our runs).
(a) If centre $j$ receives at least one point in every epoch, then $t-\tlast{j}\le 2L-2$ at every revival check, so the abandonment clause with $W=2$ never applies to $j$.
(b) Suppose, in addition, that the order of each epoch is a uniformly random permutation of the training points, independent of the orders of the other epochs. If $j$ is the nearest centre of a single training point, the gap between two consecutive visits of that point exceeds $W\in[1,2]$ epochs with probability $(2-W)^2/2$ in the limit $L\to\infty$.
\end{proposition}
\begin{proof}
(a) Epochs $e$ and $e+1$ consist of batches $eL,\dots,(e+1)L-1$ and $(e+1)L,\dots,(e+2)L-1$. If $j$ receives a point in batch $eL+b_1$ and next in batch $(e+1)L+b_2$, with $0\le b_1,b_2\le L-1$, the gap is $L+b_2-b_1\le 2L-1$. Since $\tlast{j}$ is updated before the check, any check strictly between the two batches has $t-\tlast{j}\le 2L-2$. Before the first such batch, $\tlast{j}=0$ and $t\le L-1$.

(b) Since the point is visited once per epoch, two consecutive visits occur in consecutive epochs, say $e$ and $e+1$. For $i=1,2$, let $p_i\in\{0,\dots,n-1\}$ be the position of the point in the order of epoch $e+i-1$, and $b_i=\lfloor p_i/B\rfloor$ the index, within that epoch, of the batch that contains it. As in (a), the gap between the two visits is $L+b_2-b_1$ batches, that is, $1+b_2/L-b_1/L$ epochs.

\emph{Distribution of the positions.} A uniformly random permutation places a given point at each of the $n$ positions with probability $1/n$, so $p_i$ is uniformly distributed on $\{0,\dots,n-1\}$. Since the orders of the two epochs are independent, so are $p_1$ and $p_2$, and hence $b_1$ and $b_2$. Moreover, $b_i/L$ differs from $p_i/n$ by less than $2/(L-1)$, because $L-1<n/B\le L$, and $p_i/n$ is uniform on $\{0,1/n,\dots,(n-1)/n\}$. Therefore, as $L\to\infty$ (which requires $n\to\infty$), the pair $(b_1/L,\,b_2/L)$ converges in distribution to a pair $(u_1,u_2)$ of independent random variables, each uniformly distributed on $[0,1]$, and the gap converges in distribution to $1+u_2-u_1$ epochs.

\emph{Probability of a long gap.} The gap $1+u_2-u_1$ exceeds $W$ if and only if $u_2-u_1>w$, where $w:=W-1\in[0,1]$. The set $\{(u_1,u_2)\in[0,1]^2: u_2-u_1>w\}$ is a right triangle whose two legs have length $1-w$, and $(u_1,u_2)$ is uniformly distributed on the unit square, so $\Pr[u_2-u_1>w]=(1-w)^2/2=(2-W)^2/2$. The distribution of $u_2-u_1$ has no atoms, so this is also the limit of the probability that the gap of the finite run exceeds $W$ epochs.
\end{proof}
Hence, with $W=2$, the abandonment clause \textsl{never removes a centre that receives at least one point in every epoch}, for any data, and no smaller $W$ has this property for every value of $n/B$ (the largest possible value of $t-\tlast{j}$ is $2L-2$). With $W=1$, the clause wrongly removes the centre of a single point between two given consecutive visits with probability approaching $1/2$ when an epoch has many batches, and at least once in a run of $T$ epochs with probability approaching $1-1/T!$ ($5/6$ for $T=3$). If the number of epochs is not an integer, e-means adds an incomplete epoch, which visits only a prefix of its random order, just before the last epoch, and the guarantee then requires the waiting period to be counted in complete epochs. The clause can only act in runs longer than $W$ epochs. Its purpose is to release a centre that has lost all its points while being far from every other centre, which the distance test alone would keep until the end of training; as noted above, our experiments show no measurable benefit from it (\cref{sec:exp-sens}).

\paragraph{Cost and implementation.} The rule needs two additional arrays of length $k$ ($D$ and $t^{\mathrm{last}}$); our implementation also keeps per-batch buffers, and two more arrays for the utility-estimate variant. $D_j$ is a per-cluster sum of the squared distances that the assignment step already returns, and $\delta_j$ requires the distances between the $|\mathcal{C}|$ candidates and the $k$ centres every $R$ batches. Our implementation is a subclass of the library's NumPy class (\texttt{NumpyEmeans}). With $\taunear=\infty$ and no abandonment clause, it produces exactly the same centres as the library, which our unit tests verify. Porting it to the library's PyTorch implementation for GPUs requires no new GPU kernel: only the same per-cluster sum, a search for the two nearest centres of each candidate in the revival step, and, in the multi-GPU version, summing $D$ across devices together with the counts and sums. The GPU implementation records each training step as a CUDA graph, a fixed sequence of GPU operations that is replayed for every batch, so the new state must be included in that recorded sequence. We leave this engineering to future work.

\section{Experiments}\label{sec:experiments}

\paragraph{Protocol.} We wrote down the experimental protocol (datasets, configurations, metrics and statistical tests) and froze it before any of the runs reported here. Three dated amendments were each written before the runs they describe. The first added random seeds for \cref{tab:main}, two more held-out datasets and two more embedding configurations. The second and third followed two rounds of independent review and corrected the utility-estimate variant (described below); all its runs were repeated after each correction, and after the third correction nearly all of them had exactly the same train and test MSE as before. The protocol and its amendments are distributed with the code. The values $\taunear=4$ and $W=2$ come from development runs on Covertype, on the synthetic far groups (\cref{app:synth}) and on 200{,}000-point subsets of DEEP and SIFT; they were not changed afterwards. All runs use the NumPy implementation of e-means in FAISS~1.16 on CPU, with nearest-centre search in single precision, and with the library's default settings (\cref{sec:background}): the scaling laws, \texttt{batch\_law="auto"} and the last-epoch mean. The methods differ only in the revival rule, and the size threshold is $\taucount=0.1$ in every dimension; we do not apply the library's default of disabling revival for $d\le2$, except in the method \Off.

\paragraph{Methods.} We compare the following revival rules; the names in parentheses are those used in the released code and records.
\begin{itemize}
\item \Count{} (\texttt{count}): count-based revival as in the library (the library's default for $d>2$).
\item \Off{} (\texttt{off}): no revival (the library's default for $d\le2$ since version 1.16).
\item \Dist{} (\texttt{near}): our rule, \cref{eq:rule}.
\item \Est{} (\texttt{exact}): the same rule with an online estimate of Fritzke's utility per point, $G_j/\tilde N_j$, in place of $\delta_j$. Here $G_j$ is an EMA of $\sum(d_2^2-d_1^2)$ over the batch points assigned to $j$, and $\tilde N_j$ is the EMA of their number with the same step sizes; both restart when a split creates the centre. As in \cref{sec:method}, $d_1$ and $d_2$ are the distances from a point to its assigned centre and to its nearest other centre. They are computed by a separate search for the two nearest centres, so the library's assignment is unchanged. A centre that has received no point since its creation has unknown utility and is not removed by the utility test; the abandonment clause still applies to it.
\item \Dist{} without abandonment: \Dist{} without the abandonment clause.
\item Lloyd@25: FAISS's implementation of Lloyd's algorithm, run for 25 iterations on all training points, the reference of the e-means paper.
\end{itemize}
For a given seed, all methods start from the same centres: either $k$ training points drawn at random (random initialization) or the centres chosen by greedy k-means++~\citep{arthur2007kmeanspp,pedregosa2011sklearn}, which selects each new centre as the best of several candidate points, each sampled with probability proportional to its squared distance to the nearest centre already selected.

\paragraph{Data.} We use three groups of datasets.
\begin{itemize}
\item \emph{Held-out datasets, never used during development:}
  \begin{itemize}
  \item KDD Cup 1999~\citep{kddcup99}: the full data, with exact duplicates removed as in NSL-KDD~\citep{tavallaee2009nslkdd}; 1.05M points with 122 dimensions after one-hot encoding of the categorical features.
  \item US Census 1990~\citep{census1990}: 2.0M unique points with 68 attributes.
  \item YearPredictionMSD~\citep{bertin2011msd}: 495k points with 90 features.
  \item GeoNames~\citep{geonames}: latitude and longitude of 1M places.
  \item Individual household electric power consumption~\citep{uci_power}: 1.89M points with 7 features.
  \item Statlog Shuttle~\citep{uci_shuttle}: 38k points with 9 features.
  \item SPACEV~1M~\citep{simhadri2022bigann}, an embedding dataset from the e-means paper's benchmarks, with $k=16{,}384$.
  \end{itemize}
\item \emph{Development datasets at larger scale:} DEEP~1M~\citep{babenko2016deep} and SIFT~1M~\citep{jegou2011bigann} with $k=16{,}384$, and DEEP~10M with $k=65{,}536$, all in the e-means paper's configurations. During development we used only the first 200{,}000 points of the same 1M-point training sets.
\item \emph{Development datasets:} Covertype~\citep{blackard1999covertype} (561k points, 54 dimensions), standardized or rescaled to $[0,1]$ per feature (min-max scaling), and the synthetic far groups (\cref{app:synth}).
\end{itemize}
In the tabular datasets we remove exact duplicates and standardize every feature (except in the min-max variant of Covertype); GeoNames is kept in degrees. As in the e-means paper, we hold out 20{,}000 points and use half of them as the test set. The numbers of epochs are $T\in\{1,2,3,5,10\}$ ($\{1,2,5,10\}$ for household power and Shuttle), and the numbers of centres are $k\in\{256,1024,4096\}$ ($\{256,1024\}$ for household power and Shuttle). We run 10 seeds per configuration, and 30 for the configuration of \cref{tab:main}. The test MSE is the MSE of the test points; as in the e-means paper, the nearest centre is found in single precision and the squared distance to it is then computed in double precision. The train MSE is the MSE of all training points (of a fixed subset of 200{,}000 points for the embedding datasets). \cref{app:protocol} gives further details of the data, the splits, the software and the hardware.

\paragraph{Statistics.} Comparisons are paired by seed. We report the median over seeds of the relative change in test MSE ($\Delta$; negative values favour \Dist), the numbers of seeds in which \Dist{} wins, ties or loses (W/T/L; a tie is $|\Delta|<10^{-4}$), and two-sided Wilcoxon signed-rank tests~\citep{wilcoxon1945}. As the protocol specifies, we apply Holm's correction for multiple comparisons~\citep{holm1979} within one family of tests per experiment: all comparisons of \Dist{} with \Count{} and with \Off{} on the held-out datasets, over both initializations, all numbers of epochs and all values of $k$ (\HolmFamilySizeHeld{} tests). The development datasets form a second family. Lloyd@25 is a fixed reference: we divide each method's median test MSE by the median test MSE of Lloyd@25 over seeds 0--9 with the same type of initialization (seeds 0--2 for the embedding datasets). These are ratios of medians, not paired ratios. \Cref{app:full} reports every configuration (\crefrange{tab:random-T1}{tab:synth}); the complete per-run records, including those of \Dist{} without abandonment, are released with the code.

\subsection{No Change Where Revival Helps}\label{sec:exp-emb}
On the embedding datasets of the e-means paper, count-based revival removes thousands of centres per run, and \textsl{the distance test never prevents a removal} (\cref{tab:emb}). Since the test keeps no candidate in any run, every removal, and hence every run, is the same as in the library: train and test MSE equal those of the library exactly in all \EmbRuns{} runs, and our unit tests verify that the centres are identical bit for bit when no candidate is kept. These runs include DEEP~10M with $k=65{,}536$, which we ran for 2 epochs with 3 seeds only to check this identity. In the library's runs on DEEP and SIFT, which have more than half a million candidates in total (a candidate is counted once at every revival check), every ratio $\delta_j/\mhat$ is at most about 2.5, and very few exceed 2 (\cref{fig:ratio}), well below the threshold $\taunear=4$. Revival is beneficial on these datasets: disabling it raises the test MSE by a few percent, in every seed and for every number of epochs (\cref{tab:emb}), which is consistent with the ablation in the e-means paper. Our setup also reproduces the e-means paper's results closely: after three epochs on DEEP~1M, e-means ends about 0.3\% below the test MSE of Lloyd@25, both in our runs and in the e-means paper (ratios 0.9968 and $0.38788/0.38926=0.9965$~\citep[Table~5]{bessa2026emeans}).

\begin{table}[t]\centering\small
\caption{Embedding datasets in the e-means paper's configurations (random initialization, 10 seeds per row; DEEP~10M: 3). Ratios use the median test MSE of Lloyd@25 over seeds 0--2; DEEP~10M has no Lloyd@25 reference. Identical runs: runs whose train and test MSE equal those of the library exactly. Splits per run: median number of removals, each followed by a split.}
\label{tab:emb}
\begin{tabular}{lcccccc}
\toprule
 & & & \multicolumn{3}{c}{Test MSE / Lloyd@25} & \\ \cmidrule(lr){4-6}
Dataset & Epochs & Identical runs & \textsc{count} & \textsc{off} & \textsc{distance} & Splits per run\\
\midrule
DEEP 1M & 1 & 10/10 & 1.0115 & 1.0420 & 1.0115 & 9010\\
DEEP 1M & 2 & 10/10 & 0.9989 & 1.0315 & 0.9989 & 9726\\
DEEP 1M & 3 & 10/10 & 0.9968 & 1.0180 & 0.9968 & 6892\\
DEEP 1M & 5 & 10/10 & 0.9938 & 1.0161 & 0.9938 & 7454\\
SIFT 1M & 1 & 10/10 & 1.0091 & 1.0330 & 1.0091 & 6902\\
SIFT 1M & 2 & 10/10 & 0.9967 & 1.0216 & 0.9967 & 7290\\
SIFT 1M & 3 & 10/10 & 0.9954 & 1.0108 & 0.9954 & 4622\\
SIFT 1M & 5 & 10/10 & 0.9928 & 1.0073 & 0.9928 & 4826\\
SPACEV 1M & 2 & 10/10 & 1.0009 & 1.0385 & 1.0009 & 14212\\
SPACEV 1M & 5 & 10/10 & 0.9917 & 1.0179 & 0.9917 & 11208\\
DEEP 10M & 2 & 3/3 & -- & -- & -- & 16451\\
\bottomrule
\end{tabular}

\end{table}

\subsection{Large Gains on Data with Isolated Groups}\label{sec:exp-main}
\cref{tab:main} reports the test MSE after two epochs from random initialization, as ratios to Lloyd@25. Relative to \Count, \Dist{} reduces the test error by large margins on the held-out datasets with isolated groups: by roughly 25--80\% on KDD Cup 99, 20--75\% on Shuttle and 10--25\% on GeoNames, depending on $k$. On Covertype, a development dataset, it reduces the error by up to about half. On household power, which, like Shuttle, was chosen before any result on it was seen, it reduces the error by a few percent. On Census 1990 and YearPredictionMSD, it changes the test error by at most a few tenths of a percent.

Relative to \Off, the median change in test error of \Dist{} is below 1.5\% in every held-out configuration of \cref{tab:main}. \textsl{At two epochs, \Dist{} therefore performs about as well as the better of the library's two settings.} On KDD Cup 99 its median test error is lower than both for all three values of $k$; against \Off, the difference is significant after correction only for $k=1024$.

\begin{figure}[t]
\centering
\includegraphics[width=\linewidth]{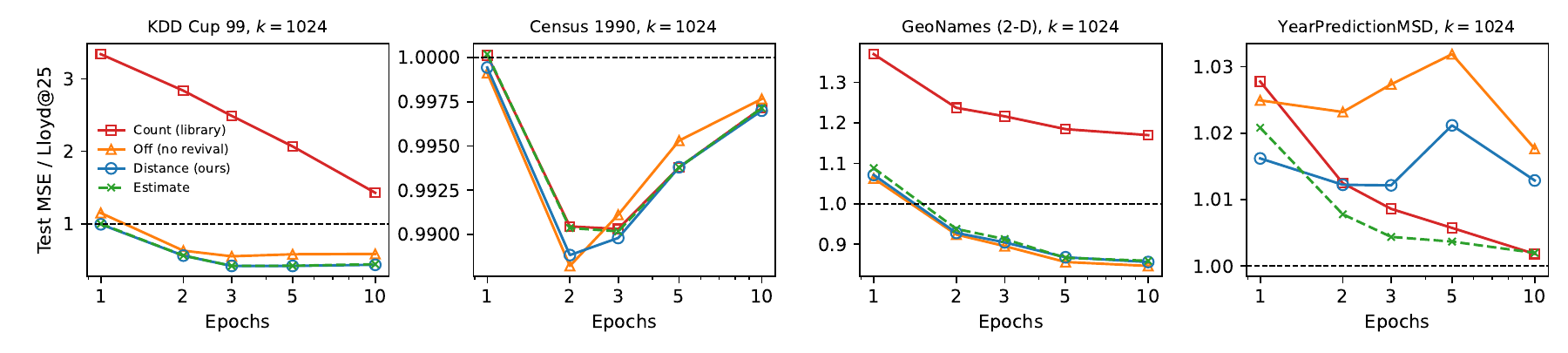}
\caption{Median test MSE divided by the median test MSE of Lloyd@25 (dashed line), as a function of the number of epochs ($k=1024$, random initialization). Each number of epochs is a separately trained run. Where curves coincide, they are drawn on top of each other: on KDD Cup 99, the curve of \Dist{} (blue circles) lies under that of \Est{} (green crosses, dashed line), from which it differs by at most 2\%; on Census 1990, the curve of \Count{} (red squares) nearly coincides with that of \Est.}
\label{fig:curves}
\end{figure}

With more epochs, \Count{} does not catch up (\cref{fig:curves}). In \NeverReachCount{} held-out configurations, \Count{} does not reach the test error of Lloyd@25 at any tested number of epochs up to ten (\cref{tab:epochs-random}). \Dist{} reaches it in \NearReachCount{} of these configurations, and within two epochs in \NearReachWithinTwo{} of them. Here, the number of epochs needed is the smallest tested number of epochs whose median test error is below that of Lloyd@25; each number of epochs is a separate run. The two remaining configurations are on YearPredictionMSD, where neither method reaches it. Outside the embedding setting, the distance test therefore restores, in most configurations, the main property of e-means: reaching the error of Lloyd@25 in a few epochs. On GeoNames, the test keeps almost every candidate, and \Dist{} performs like \Off, which is the library's default for $d\le2$ since version 1.16, without a rule based on dimension.

\begin{table}[t]\centering\small
\caption{Test MSE after two epochs, random initialization. Columns 4--7: median test MSE divided by the median test MSE of Lloyd@25 over seeds 0--9 (a fixed reference, not a paired ratio). Seeds: number of paired seeds per comparison (30, or 10 for household power and Shuttle). $\Delta$: median over seeds of the relative change in the test MSE of \Dist{} with respect to the other method (negative values favour \Dist). W/T/L: numbers of seeds in which \Dist{} is better, tied ($|\Delta|<10^{-4}$) or worse. $p$: two-sided Wilcoxon signed-rank test, Holm-adjusted over all comparisons on the held-out datasets (the development datasets form a separate family). \Cref{tab:random-T1,tab:random-T3,tab:random-T5,tab:random-T10} in the appendix report the other numbers of epochs.}
\label{tab:main}
\resizebox{\linewidth}{!}{\begin{tabular}{lllcccccccccc}
\toprule
 & & & \multicolumn{4}{c}{Median test MSE / median test MSE of Lloyd@25} & \multicolumn{3}{c}{\textsc{distance} vs.\ \textsc{count}} & \multicolumn{3}{c}{\textsc{distance} vs.\ \textsc{off}}\\
\cmidrule(lr){4-7}\cmidrule(lr){8-10}\cmidrule(lr){11-13}
Dataset & $k$ & Seeds & \textsc{count} & \textsc{off} & \textsc{distance} & \textsc{estimate} & $\Delta$ & W/T/L & $p$ & $\Delta$ & W/T/L & $p$\\
\midrule
\multicolumn{13}{l}{\emph{Held-out datasets}}\\
KDD Cup 99 & 256 & 30 & 1.121 & 0.550 & \textbf{0.526} & 0.524 & $-$51.8\% & 30/0/0 & $<10^{-6}$ & $-$3.1\% & 22/0/8 & 1.000 \\
KDD Cup 99 & 1024 & 30 & 2.837 & 0.629 & \textbf{0.561} & 0.562 & $-$81.2\% & 30/0/0 & $<10^{-6}$ & $-$8.7\% & 27/0/3 & 0.007 \\
KDD Cup 99 & 4096 & 30 & 1.174 & 0.982 & \textbf{0.916} & 0.960 & $-$23.6\% & 26/0/4 & $<10^{-3}$ & $-$5.6\% & 23/0/7 & 0.108 \\
Census 1990 & 256 & 30 & 0.991 & 0.992 & \textbf{0.991} & 0.991 & +0.0\% & 3/19/8 & 1.000 & $-$0.0\% & 16/2/12 & 1.000 \\
Census 1990 & 1024 & 30 & 0.990 & 0.988 & \textbf{0.989} & 0.990 & $-$0.1\% & 22/4/4 & 0.006 & +0.0\% & 12/0/18 & 1.000 \\
Census 1990 & 4096 & 30 & 0.992 & 0.990 & \textbf{0.991} & 0.992 & $-$0.1\% & 24/1/5 & 0.004 & +0.0\% & 12/1/17 & 1.000 \\
YearPredictionMSD & 256 & 30 & 1.020 & 1.025 & \textbf{1.021} & 1.011 & +0.1\% & 10/0/20 & 1.000 & $-$0.5\% & 29/0/1 & $<10^{-6}$ \\
YearPredictionMSD & 1024 & 30 & 1.012 & 1.023 & \textbf{1.012} & 1.008 & +0.0\% & 14/0/16 & 1.000 & $-$1.1\% & 30/0/0 & $<10^{-6}$ \\
YearPredictionMSD & 4096 & 30 & 1.003 & 1.029 & \textbf{1.005} & 1.003 & +0.2\% & 1/0/29 & $<10^{-5}$ & $-$2.3\% & 30/0/0 & $<10^{-6}$ \\
GeoNames (2-D) & 256 & 30 & 1.027 & 0.902 & \textbf{0.903} & 0.910 & $-$9.2\% & 29/0/1 & $<10^{-6}$ & +0.0\% & 12/11/7 & 1.000 \\
GeoNames (2-D) & 1024 & 30 & 1.237 & 0.924 & \textbf{0.928} & 0.938 & $-$25.4\% & 30/0/0 & $<10^{-6}$ & $-$0.2\% & 20/0/10 & 1.000 \\
GeoNames (2-D) & 4096 & 30 & 1.120 & 0.963 & \textbf{0.939} & 0.964 & $-$14.0\% & 29/0/1 & $<10^{-6}$ & $-$1.4\% & 23/0/7 & 0.504 \\
Household power & 256 & 10 & 1.003 & 0.945 & \textbf{0.949} & 0.953 & $-$5.1\% & 10/0/0 & 0.504 & +0.5\% & 4/0/6 & 1.000 \\
Household power & 1024 & 10 & 0.972 & 0.949 & \textbf{0.960} & 0.967 & $-$1.5\% & 10/0/0 & 0.504 & +0.7\% & 0/0/10 & 0.504 \\
Shuttle & 256 & 10 & 3.853 & 1.021 & \textbf{1.041} & 1.063 & $-$73.3\% & 10/0/0 & 0.504 & +1.3\% & 1/0/9 & 1.000 \\
Shuttle & 1024 & 10 & 1.350 & 1.038 & \textbf{1.039} & 1.037 & $-$20.2\% & 10/0/0 & 0.504 & +0.9\% & 2/0/8 & 1.000 \\
\midrule
\multicolumn{13}{l}{\emph{Development datasets}}\\
Covertype (z-score) & 256 & 30 & 1.576 & 0.757 & \textbf{0.747} & 0.746 & $-$52.5\% & 30/0/0 & $<10^{-6}$ & $-$1.0\% & 28/0/2 & $<10^{-5}$ \\
Covertype (z-score) & 1024 & 30 & 1.002 & 1.005 & \textbf{0.996} & 0.997 & $-$0.3\% & 23/1/6 & 0.109 & $-$0.7\% & 30/0/0 & $<10^{-6}$ \\
Covertype (z-score) & 4096 & 30 & 1.008 & 1.008 & \textbf{1.008} & 1.007 & $-$0.1\% & 21/1/8 & 0.819 & $-$0.1\% & 19/0/11 & 1.000 \\
Covertype (min-max) & 256 & 30 & 0.980 & 0.973 & \textbf{0.973} & 0.975 & $-$0.6\% & 24/0/6 & 0.605 & $-$0.0\% & 18/2/10 & 1.000 \\
Covertype (min-max) & 1024 & 30 & 1.011 & 1.006 & \textbf{1.003} & 1.006 & $-$0.4\% & 24/0/6 & 0.101 & $-$0.3\% & 24/0/6 & 0.223 \\
Covertype (min-max) & 4096 & 30 & 1.003 & 1.007 & \textbf{1.003} & 1.003 & $-$0.1\% & 20/2/8 & 1.000 & $-$0.2\% & 25/0/5 & 0.018 \\
\bottomrule
\end{tabular}
}
\end{table}

\subsection{k-means++ Initialization}\label{sec:exp-kmpp}
With k-means++ initialization, the failure of count-based revival is larger: k-means++ places centres on rare groups, and count-based revival removes them. After two epochs with $k=1024$, \Count{} ends far above the test error of Lloyd@25 started from the same centres: at more than 25 times that error on KDD Cup 99, and 76\% above it on GeoNames. \Dist{} keeps those centres and ends within 1\% of Lloyd@25 on both datasets, as does \Off{} (\cref{tab:kmpp-T2}; \cref{app:kmpp} gives the other numbers of epochs).

\subsection{Bound versus Utility Estimate}\label{sec:exp-exact}
In most configurations, the online utility estimate (\Est) is not better than the bound (\Dist); YearPredictionMSD is the exception (\cref{tab:exact}).

\paragraph{Rare points far from all others.} On such data, the bound is much better in training error. On KDD Cup 99 after 2 epochs ($k\ge1024$), the training error of \Dist{} is roughly 30--40\% lower than that of \Est, and its test error is slightly lower (by less than 2\%). Covertype shows the same pattern in training error.

\paragraph{Other datasets.} On Census, GeoNames, Shuttle and the embedding datasets, the two are equal or the bound is slightly better. On household power, the bound is better after 2 epochs and the estimate after 5 epochs, by less than 1\% in both cases. On YearPredictionMSD, the estimate is better: its test MSE is up to about 2\% lower than that of \Dist{} (\ExactBetterMsdWhere).

\paragraph{k-means++ initialization.} Because centres of unknown utility are kept (Amendment~2 of the protocol), \Est{} also keeps the centres placed by k-means++, and ends at about the test error of Lloyd@25 on KDD Cup 99 and GeoNames with $k=1024$. Under the original convention, which treated the removal of a centre that had not yet received any point as free, it removed them.

\paragraph{Running time.} The estimate requires an additional search for the two nearest centres of every point in every batch, which makes runs markedly slower (\cref{sec:exp-cost}). On GPUs, the library's assignment kernel, which returns only the nearest centre, would need a variant that also returns the second-nearest one.

We keep the bound as the default: in our timings it adds no measurable running time, and it requires only the centres, not the points of the cluster.

\subsection{Sensitivity and the Abandonment Clause}\label{sec:exp-sens}

\paragraph{Threshold $\taunear$} (\cref{tab:sens}; 10 seeds). On KDD Cup 99 and GeoNames, values from 2 to 8 give nearly the same results (within about 1.5 percentage points of each other). Larger values keep fewer candidates and give up part of the gain, up to several percentage points for $\taunear=32$ on KDD Cup 99 after 10 epochs. On Covertype with $k=1024$, every value stays within 1\% of \Count, since few candidates there are far from other centres. On DEEP~1M, $\taunear=1$ keeps redundant candidates and raises the test MSE by about 2\%; $\taunear=2$ keeps several hundred candidates in total over ten runs, with a negligible change in test MSE; and with $\taunear\ge4$, the runs are identical to those of the library. \textsl{In our experiments, the exact value of $\taunear$ therefore matters little, as long as it lies roughly between 2 and 8.}

\paragraph{Abandonment clause.} The clause can act only in runs longer than two epochs. In those runs on the held-out datasets, it removes at least one candidate in about 9\% of the runs, and it accounts for about 0.2\% of all removals (\AbandonRunsHit{} of \AbandonRuns{} runs; \AbandonRemovals{} of \AbandonAllRemovals{} removals). Removing it changes the median test MSE negligibly (by at most a few hundredths of a percent) in every held-out configuration. In our experiments, the abandonment clause therefore had no measurable benefit, and the gains come from the distance test. We keep it as the safeguard motivated in \cref{sec:method}, since it adds no measurable cost; a simpler rule could omit it.

\paragraph{Waiting period.} Shortening it to $W=1$ epoch is harmful, as \cref{prop:window} predicts. On KDD Cup 99 ($k=1024$) after 5 epochs, the reduction in training error relative to \Count{} falls from about 80\% to about 65\%.

\subsection{Running Time}\label{sec:exp-cost}
We timed the library, \Dist{} and \Est{} on one server, with an AMD Ryzen Threadripper 3990X processor (64 cores), 188~GiB of memory and Ubuntu~24.04, using 16 threads and with no other jobs of ours running (\cref{app:protocol} gives the full specification and the software versions), for 2 epochs from the same random initialization, on DEEP~1M ($k=16{,}384$) and KDD Cup 99 ($k=4{,}096$). We repeated each measurement \TimingReps{} times and rotated the order of the three methods across repetitions (\cref{tab:time}). In both configurations, \textsl{the additional work of the distance test (one EMA, and a nearest-centre search for the candidates every five batches) is within the variation between repetitions}: per repetition, the running time of \Dist{} is within a few percent of that of the library, in both directions. The search for the two nearest centres of every point in every batch makes \Est{} more than twice as slow with FAISS's exact (brute-force) CPU search.

\begin{table}[t]\centering\small
\caption{Running time in seconds, 2 epochs from random initialization, one server (AMD Ryzen Threadripper 3990X) with 16 threads; medians over \TimingReps{} repetitions. Per repetition, the ratio of the running time of \Dist{} to that of the library is \NearRatioMin--\NearRatioMax.}
\label{tab:time}
\begin{tabular}{lccc}
\toprule
Dataset & Library (\Count) & \Dist & \Est\\
\midrule
DEEP~1M, $k=16{,}384$ & \TimeDeepLib & \TimeDeepNear & \TimeDeepExact\\
KDD Cup 99, $k=4{,}096$ & \TimeKddLib & \TimeKddNear & \TimeKddExact\\
\bottomrule
\end{tabular}
\end{table}

\subsection{Mechanism and Downsides}\label{sec:exp-mech}

\begin{figure}[t]
\centering
\includegraphics[width=0.62\linewidth]{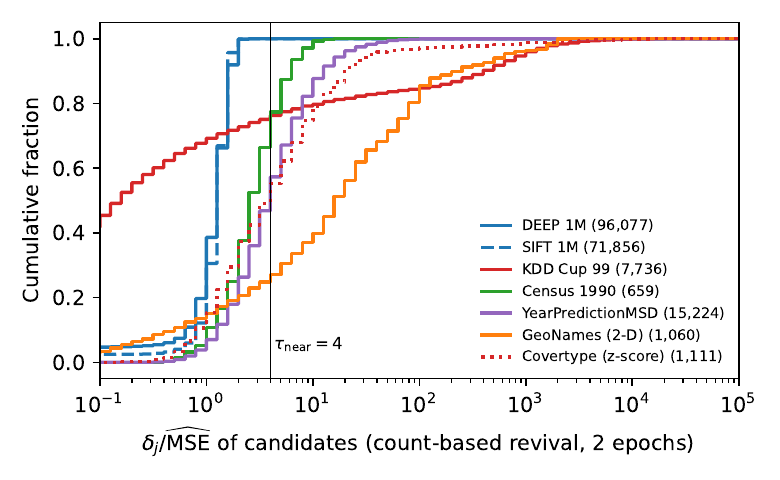}
\caption{Cumulative distribution of the ratio $\delta_j/\mhat$ that the distance test computes for each candidate, recorded in runs of count-based revival, which remove every candidate whatever its ratio (2 epochs, random initialization; $k=1024$, or $k=16{,}384$ for DEEP~1M and SIFT~1M). Each curve pools all the runs of one dataset (30 seeds for the tabular datasets, 10 for DEEP~1M and SIFT~1M), and counts a candidate once at every revival check at which it is a candidate; the legend gives the number of pooled values. A curve is therefore the distribution of these pooled values, not an average over runs. The values are recorded in a histogram with ten bins per factor of ten, hence the steps. No value on DEEP~1M or SIFT~1M exceeds about 2.5. The fraction above $\taunear=4$ is \FracAboveFourKdd\% on KDD Cup 99, \FracAboveFourCensus\% on Census, \FracAboveFourMsd\% on YearPredictionMSD, \FracAboveFourCov\% on Covertype and \FracAboveFourGeo\% on GeoNames; each range reflects the width of the histogram bin that contains 4. On KDD Cup 99 and Covertype, part of the distribution lies several orders of magnitude above the threshold.}
\label{fig:ratio}
\end{figure}

\paragraph{Mechanism.} \cref{fig:ratio} shows why a single threshold separates the two settings. On the embedding datasets, every candidate is close to another centre. On the tabular datasets, a minority of candidates, whose removal is very costly, lie far from all other centres (up to five orders of magnitude above $\mhat$ on KDD Cup 99 and Covertype), and on GeoNames most candidates do. The gains come from the points with the largest errors: the mean of the largest 0.1\% of the per-point training errors decreases more than fourfold on KDD Cup 99 ($k=1024$) and by almost two orders of magnitude on Covertype ($k=256$) (\cref{tab:imbalance}). On the synthetic far groups, \Count{} keeps no far group of up to 150 points (less than a tenth of the size of a typical cluster), whereas \Dist{} keeps about as many as \Off{} and Lloyd@25, and all five from k-means++ initialization once a group has at least 150 points (\cref{tab:synth}).

\paragraph{Downsides.} We observed two, both anticipated by the analysis.

The first is that the test ignores the error reduction of the split, which shows on YearPredictionMSD. There, the largest increase in test error of \Dist{} relative to \Count{} is below 3\% (\MsdWorstWhere; higher in all ten seeds), and such increases appear for larger $k$ and more epochs. To understand it, we logged both quantities along one run of \Dist{} ($k=4096$, 10 epochs, seed 0). The test keeps most candidates (more than 90\%, counted once per revival check). For about four-fifths of the kept candidates, the utility estimate also exceeds $\taunear\mhat$; for these, the bound and the estimate nearly coincide (median ratio close to 1), and the median estimated utility per point is about ten times $\mhat$. Most kept clusters are therefore costly per point by either measure. A plausible explanation is that splitting large clusters reduces the error more than keeping these few costly points saves, and the per-point test does not account for that reduction. We did not measure the error reduction of the splits directly. The kept candidates on which the two measures disagree may also contribute, through their effect on the rest of the run. Keeping candidates also makes the cluster sizes less uniform on this dataset: the imbalance $k\sum_j|P_j|^2/n^2$, which equals 1 when all clusters have the same size, rises by almost half for $k=256$ (\cref{tab:imbalance}).

The second is the single, global error scale, which shows when cluster scales differ widely. On a two-dimensional mixture of 90 tight and 10 broad Gaussian blobs (\cref{app:synth}), the test keeps redundant small clusters inside the broad blobs. The median change in MSE of \Dist{} relative to \Count{} is then an increase of about 7\% with random initialization, but a decrease of about 13\% with k-means++ initialization, because \Count{} then removes the centres that k-means++ placed in sparse regions. In 16 dimensions, the median change is negligible (\cref{tab:synth}).

Both downsides point to the same refinement: compare the cost of a removal with the error reduction of the split that would replace the removed cluster, and measure both relative to a local error scale.

\section{Related Work}\label{sec:related}

\paragraph{Removal cost and utility.} The cost of removing a centre while the other centres are kept fixed, $\sum_{x}(d_2^2-d_1^2)$ over its points, is Fritzke's \textsl{utility}. Methods that use it differ in how they act on it. LBG-U and k-means-u*~\citep{fritzke1997lbgu,fritzke2017kmeansu} move the centre of lowest utility next to the centre with the largest error, run Lloyd's algorithm on the full data, and keep the move only if the total error decreases. Breathing k-means~\citep{fritzke2020breathing} alternately adds centres near centres with large error and removes the centres of lowest utility, with Lloyd's algorithm in between; it does not remove the nearest neighbour of a centre selected for removal in the same step (freezing), so that two adjacent centres are not removed together. GNG-U~\citep{fritzke1997gngu} is an online rule for growing neural gas~\citep{fritzke1995gng}, a network of units (centres) that is adapted one point at a time; it removes the unit of lowest utility when the ratio of the largest error to that utility exceeds a threshold, so removal is triggered by the utility itself, not by a count. Iterative shrinking and the pairwise nearest neighbour (PNN) method~\citep{franti2006shrinking,ward1963,equitz1989} reduce the number of clusters greedily, removing the cluster, or merging the pair of clusters, whose removal or merge costs least. Fritzke already notes that centres with close neighbours tend to have lower utility than isolated ones~\citep{fritzke2017kmeansu}. Our contribution is narrower: for mini-batch k-means, a bound on the removal cost that is computed from the centres alone and used as an additional condition on a count-based removal rule, so that the algorithm is unchanged wherever the count-based rule already works.

\paragraph{Revival in mini-batch k-means and vector-quantized neural networks.} All the revival rules for mini-batch k-means and for vector-quantized neural networks that we found decide by count, or by the time since a centre was last used, alone. They include the reassignment step of scikit-learn's \texttt{MiniBatchKMeans}~\citep{sculley2010,sklearn_minibatch_code}, e-means~\citep{bessa2026emeans}, and the resets of rarely used code vectors (the learned centres) in neural vector quantizers~\citep{dhariwal2020jukebox,zeghidour2021soundstream,lucidrains_vq,huh2023straightening,zheng2023cvq}; one of these, the vector-quantize-pytorch library, uses the same threshold as e-means (a tenth of the average count). Full-batch implementations relocate only empty clusters~\citep{douze2024faiss,pedregosa2011sklearn}. ISODATA~\citep{ball1965isodata} discards small clusters and merges close ones, but through two separate rules. Online k-means with remove-birth updating~\citep{fujita2023okrb} moves the unit that has been nearest to the fewest points, again a count-based criterion. We are not aware of a count-based revival rule that also checks whether the centre of a small cluster is far from the other centres.

\paragraph{Outliers.} Outlier-aware variants of k-means, such as k-means--~\citep{chawla2013kmeansmm}, decide which \textsl{points} to ignore. Our question is which \textsl{centres} to keep: a small isolated group is what an outlier-aware method may discard, and what the k-means objective rewards covering.

\section{Limitations and Future Work}\label{sec:limitations}

\paragraph{Scale and hardware.} All results come from CPU runs of the library's NumPy implementation. The tabular datasets have at most 2M points. The embedding evaluations cover the e-means paper's configurations with $10^6$ points and one configuration with $10^7$ points (3 seeds, 2 epochs), in which we only verified that the results are identical to the library's. We did not run the GPU or multi-GPU implementations, or datasets with more than $10^8$ points. The rule requires no new GPU kernel (\cref{sec:method}), but its running time in the GPU implementation, with CUDA-graph replay, is unmeasured.

\paragraph{Hyperparameters.} $\taunear=4$ was chosen on the development data as a margin above the largest ratio $\delta_j/\mhat$ observed on embedding data, and fixed before the evaluation on held-out datasets. On held-out data, values from 2 to 8 give nearly the same results, and larger values give up part of the gain (\cref{sec:exp-sens}). A principled choice, for example one that compares the removal cost with the expected error reduction of the split that would replace the removed cluster, remains open.

\paragraph{Simultaneous removals.} The bound concerns the removal of one centre at a time; two neighbouring candidates can be removed together (\cref{sec:method}). Freezing, as in breathing k-means, would prevent this; we did not measure how often it occurs.

\paragraph{A single error scale.} $\mhat$ is an average over all clusters. When cluster scales differ widely, a redundant small cluster in a broad region can appear far from other centres relative to $\mhat$, and be kept. Our heterogeneous-scale experiment shows this effect in two dimensions (a median increase in MSE of about 7\% relative to count-based revival, from random initialization) but not in sixteen (\cref{sec:exp-mech}). A local scale, for example the per-point error of the neighbouring cluster, is a natural variant.

\paragraph{Downsides.} Keeping small clusters spends centres on rare groups, which the k-means objective rewards, but makes cluster sizes less uniform where many candidates are kept (on YearPredictionMSD with $k=256$, the imbalance rises by almost half); elsewhere the change is small (\cref{tab:imbalance}). The same dataset shows the cost of ignoring the error reduction of the split (\cref{sec:exp-mech}). For an inverted-file index for nearest-neighbour search, which stores one list of points per cluster~\citep{douze2024faiss}, balanced cluster sizes may matter more than squared error. On the embedding datasets we tested, the rule changes nothing.

\paragraph{Running time.} Our running-time measurements cover two configurations on one CPU server.

\paragraph{Future work.} GPU and multi-GPU implementations; comparing the removal cost with the error reduction of the split that would replace the removed cluster; using the same per-point quantity to choose donors (splitting where the error reduction is largest) and split directions; and applying the test to the count-based resets of code vectors in vector-quantized neural networks.

\section*{AI Use Statement}

Claude Opus~5.5 (Anthropic), used through Claude Code, was used for coding, experiments and writing. GPT-6.1-sol (OpenAI), used through Codex, was used for paper and code reviews, and Claude Fable~5.2 (Anthropic) for reviewing the theoretical results. The authors reviewed all AI-assisted text, figures, and code. We take responsibility for the final content of this paper, including all claims.

\bibliographystyle{plainnat}
\bibliography{refs}

\appendix
\crefalias{section}{appendix}
\crefalias{subsection}{subappendix}
\section{Full Results}\label{app:full}

This appendix gives the full results behind \cref{sec:experiments}, in \crefrange{tab:random-T1}{tab:synth}: the non-embedding datasets from random initialization (\cref{app:random}) and from k-means++ initialization (\cref{app:kmpp}), the comparison of the bound with the utility estimate (\cref{app:exact}), the sensitivity to the hyperparameters (\cref{app:sens}), the cluster-size imbalance and the tail error (\cref{app:imbalance}), and the synthetic data (\cref{app:synth}). \Cref{app:protocol} gives further details of the data and of the protocol. The following conventions apply to all these tables.

\paragraph{Seeds.} Most configurations have 10 seeds (0--9). The configuration with random initialization and 2 epochs has 30 seeds (0--29) for the six original tabular datasets. The comparisons with \Est{} on the embedding datasets have 5 seeds, and the DEEP~10M check has 3. The column ``Seeds'' gives the number of paired seeds behind each comparison.

\paragraph{Notation.} W/T/L counts the seeds in which the test MSE of \Dist{} is lower than, within a relative difference of $10^{-4}$ of, or higher than that of the other method. $\Delta$ is the median over seeds of the relative change in test MSE. $p$ is the $p$-value of a two-sided Wilcoxon signed-rank test~\citep{wilcoxon1945}, Holm-adjusted~\citep{holm1979} within one family of tests per experiment: all \HolmFamilySizeHeld{} comparisons on the held-out datasets, and separately the \HolmFamilySizeDev{} comparisons on the development datasets.

\paragraph{Lloyd reference.} The ratio columns divide each method's median test MSE by the median test MSE of Lloyd@25 (FAISS's implementation of Lloyd's algorithm, 25 iterations on all training points) over seeds 0--9 with the same type of initialization (seeds 0--2 for the embedding datasets; DEEP~10M has no reference). For a given seed, Lloyd@25 and the e-means methods start from the same centres.

\paragraph{Released records.} The runs of \Dist{} without the abandonment clause, all timings and all per-run metrics are in the released records (\texttt{results/runs.jsonl.gz}).

\subsection{Non-Embedding Datasets, Random Initialization}\label{app:random}
\Cref{tab:random-T1,tab:random-T3,tab:random-T5,tab:random-T10} complement \cref{tab:main}, which reports 2 epochs, with 1, 3, 5 and 10 epochs; their columns are those of \cref{tab:main}. \Cref{tab:epochs-random} gives, for each configuration, the smallest tested number of epochs at which each method's median test MSE is below that of Lloyd@25, which \cref{sec:exp-main} summarizes.

\begin{table}[H]\centering\small
\caption{Random initialization, 1 epoch. Columns as in \cref{tab:main}.}\label{tab:random-T1}\resizebox{\linewidth}{!}{\begin{tabular}{lllcccccccccc}
\toprule
 & & & \multicolumn{4}{c}{Median test MSE / median test MSE of Lloyd@25} & \multicolumn{3}{c}{\textsc{distance} vs.\ \textsc{count}} & \multicolumn{3}{c}{\textsc{distance} vs.\ \textsc{off}}\\
\cmidrule(lr){4-7}\cmidrule(lr){8-10}\cmidrule(lr){11-13}
Dataset & $k$ & Seeds & \textsc{count} & \textsc{off} & \textsc{distance} & \textsc{estimate} & $\Delta$ & W/T/L & $p$ & $\Delta$ & W/T/L & $p$\\
\midrule
\multicolumn{13}{l}{\emph{Held-out datasets}}\\
KDD Cup 99 & 256 & 10 & 1.194 & 0.812 & \textbf{0.840} & 0.853 & $-$30.4\% & 10/0/0 & 0.504 & +3.4\% & 3/0/7 & 1.000 \\
KDD Cup 99 & 1024 & 10 & 3.341 & 1.150 & \textbf{0.993} & 1.000 & $-$70.8\% & 10/0/0 & 0.504 & $-$3.7\% & 6/0/4 & 1.000 \\
KDD Cup 99 & 4096 & 10 & 2.061 & 1.434 & \textbf{1.338} & 1.501 & $-$27.9\% & 10/0/0 & 0.504 & $-$2.0\% & 6/0/4 & 1.000 \\
Census 1990 & 256 & 10 & 0.997 & 0.996 & \textbf{0.998} & 0.997 & +0.0\% & 2/7/1 & 1.000 & $-$0.1\% & 7/0/3 & 1.000 \\
Census 1990 & 1024 & 10 & 1.000 & 0.999 & \textbf{0.999} & 1.000 & $-$0.0\% & 5/0/5 & 1.000 & $-$0.0\% & 5/1/4 & 1.000 \\
Census 1990 & 4096 & 10 & 1.005 & 1.002 & \textbf{1.003} & 1.004 & $-$0.1\% & 9/0/1 & 0.721 & +0.0\% & 5/0/5 & 1.000 \\
YearPredictionMSD & 256 & 10 & 1.040 & 1.018 & \textbf{1.017} & 1.025 & $-$2.2\% & 10/0/0 & 0.504 & $-$0.0\% & 5/1/4 & 1.000 \\
YearPredictionMSD & 1024 & 10 & 1.028 & 1.025 & \textbf{1.016} & 1.021 & $-$1.0\% & 10/0/0 & 0.504 & $-$0.8\% & 10/0/0 & 0.504 \\
YearPredictionMSD & 4096 & 10 & 1.017 & 1.039 & \textbf{1.014} & 1.018 & $-$0.3\% & 8/0/2 & 1.000 & $-$2.3\% & 10/0/0 & 0.504 \\
GeoNames (2-D) & 256 & 10 & 1.112 & 1.007 & \textbf{1.005} & 1.038 & $-$8.5\% & 10/0/0 & 0.504 & +0.0\% & 4/2/4 & 1.000 \\
GeoNames (2-D) & 1024 & 10 & 1.370 & 1.063 & \textbf{1.071} & 1.088 & $-$22.7\% & 10/0/0 & 0.504 & $-$0.3\% & 7/0/3 & 1.000 \\
GeoNames (2-D) & 4096 & 10 & 1.205 & 1.083 & \textbf{1.063} & 1.085 & $-$12.1\% & 10/0/0 & 0.504 & $-$1.5\% & 7/0/3 & 1.000 \\
Household power & 256 & 10 & 1.040 & 0.983 & \textbf{0.989} & 0.990 & $-$5.1\% & 10/0/0 & 0.504 & +0.5\% & 2/1/7 & 1.000 \\
Household power & 1024 & 10 & 1.025 & 1.004 & \textbf{1.006} & 1.011 & $-$2.0\% & 10/0/0 & 0.504 & +0.4\% & 2/0/8 & 1.000 \\
Shuttle & 256 & 10 & 4.452 & 2.280 & \textbf{2.765} & 3.037 & $-$37.4\% & 10/0/0 & 0.504 & +14.5\% & 1/0/9 & 0.508 \\
Shuttle & 1024 & 10 & 3.001 & 1.543 & \textbf{1.544} & 1.554 & $-$43.7\% & 10/0/0 & 0.504 & +1.1\% & 2/0/8 & 1.000 \\
\midrule
\multicolumn{13}{l}{\emph{Development datasets}}\\
Covertype (z-score) & 256 & 10 & 1.794 & 0.841 & \textbf{0.831} & 0.832 & $-$52.2\% & 10/0/0 & 0.223 & $-$1.3\% & 10/0/0 & 0.223 \\
Covertype (z-score) & 1024 & 10 & 1.043 & 1.024 & \textbf{1.023} & 1.020 & $-$0.5\% & 8/0/2 & 0.793 & $-$0.3\% & 9/0/1 & 1.000 \\
Covertype (z-score) & 4096 & 10 & 1.049 & 1.049 & \textbf{1.048} & 1.048 & $-$0.0\% & 8/0/2 & 1.000 & +0.0\% & 4/0/6 & 1.000 \\
Covertype (min-max) & 256 & 10 & 0.998 & 0.982 & \textbf{0.985} & 0.995 & $-$0.5\% & 7/0/3 & 1.000 & +0.0\% & 4/1/5 & 1.000 \\
Covertype (min-max) & 1024 & 10 & 1.019 & 1.017 & \textbf{1.007} & 1.009 & $-$0.8\% & 9/0/1 & 0.393 & $-$0.4\% & 9/0/1 & 0.285 \\
Covertype (min-max) & 4096 & 10 & 1.040 & 1.040 & \textbf{1.038} & 1.039 & $-$0.2\% & 9/1/0 & 0.285 & $-$0.0\% & 5/0/5 & 1.000 \\
\bottomrule
\end{tabular}
}
\end{table}
\begin{table}[H]\centering\small
\caption{Random initialization, 3 epochs.}\label{tab:random-T3}\resizebox{\linewidth}{!}{\begin{tabular}{lllcccccccccc}
\toprule
 & & & \multicolumn{4}{c}{Median test MSE / median test MSE of Lloyd@25} & \multicolumn{3}{c}{\textsc{distance} vs.\ \textsc{count}} & \multicolumn{3}{c}{\textsc{distance} vs.\ \textsc{off}}\\
\cmidrule(lr){4-7}\cmidrule(lr){8-10}\cmidrule(lr){11-13}
Dataset & $k$ & Seeds & \textsc{count} & \textsc{off} & \textsc{distance} & \textsc{estimate} & $\Delta$ & W/T/L & $p$ & $\Delta$ & W/T/L & $p$\\
\midrule
\multicolumn{13}{l}{\emph{Held-out datasets}}\\
KDD Cup 99 & 256 & 10 & 1.003 & 0.523 & \textbf{0.497} & 0.528 & $-$51.6\% & 10/0/0 & 0.504 & $-$7.7\% & 7/0/3 & 1.000 \\
KDD Cup 99 & 1024 & 10 & 2.489 & 0.552 & \textbf{0.419} & 0.420 & $-$82.2\% & 10/0/0 & 0.504 & $-$20.3\% & 9/0/1 & 1.000 \\
KDD Cup 99 & 4096 & 10 & 1.084 & 1.007 & \textbf{0.908} & 0.922 & $-$17.2\% & 9/0/1 & 1.000 & $-$8.0\% & 9/0/1 & 1.000 \\
Census 1990 & 256 & 10 & 0.993 & 0.997 & \textbf{0.992} & 0.993 & +0.0\% & 1/9/0 & 1.000 & $-$0.4\% & 9/0/1 & 0.508 \\
Census 1990 & 1024 & 10 & 0.990 & 0.991 & \textbf{0.990} & 0.990 & $-$0.1\% & 9/0/1 & 0.508 & $-$0.0\% & 5/1/4 & 1.000 \\
Census 1990 & 4096 & 10 & 0.990 & 0.987 & \textbf{0.989} & 0.990 & $-$0.1\% & 8/0/2 & 1.000 & +0.1\% & 3/0/7 & 1.000 \\
YearPredictionMSD & 256 & 10 & 1.016 & 1.031 & \textbf{1.026} & 1.009 & +0.8\% & 0/0/10 & 0.504 & $-$0.5\% & 10/0/0 & 0.504 \\
YearPredictionMSD & 1024 & 10 & 1.009 & 1.027 & \textbf{1.012} & 1.004 & +0.4\% & 2/0/8 & 1.000 & $-$1.5\% & 10/0/0 & 0.504 \\
YearPredictionMSD & 4096 & 10 & 1.000 & 1.021 & \textbf{1.004} & 1.001 & +0.5\% & 0/1/9 & 0.504 & $-$1.8\% & 10/0/0 & 0.504 \\
GeoNames (2-D) & 256 & 10 & 1.024 & 0.886 & \textbf{0.882} & 0.903 & $-$9.7\% & 10/0/0 & 0.504 & +0.0\% & 3/3/4 & 1.000 \\
GeoNames (2-D) & 1024 & 10 & 1.216 & 0.895 & \textbf{0.905} & 0.913 & $-$27.9\% & 9/0/1 & 0.508 & +0.1\% & 4/1/5 & 1.000 \\
GeoNames (2-D) & 4096 & 10 & 1.064 & 0.942 & \textbf{0.942} & 0.950 & $-$11.8\% & 10/0/0 & 0.504 & $-$0.6\% & 7/0/3 & 1.000 \\
\midrule
\multicolumn{13}{l}{\emph{Development datasets}}\\
Covertype (z-score) & 256 & 10 & 1.445 & 0.752 & \textbf{0.743} & 0.743 & $-$48.7\% & 10/0/0 & 0.223 & $-$1.4\% & 10/0/0 & 0.223 \\
Covertype (z-score) & 1024 & 10 & 0.985 & 0.992 & \textbf{0.983} & 0.982 & $-$0.3\% & 9/0/1 & 0.793 & $-$0.9\% & 10/0/0 & 0.223 \\
Covertype (z-score) & 4096 & 10 & 1.000 & 1.002 & \textbf{1.000} & 1.000 & $-$0.1\% & 7/0/3 & 1.000 & $-$0.1\% & 10/0/0 & 0.223 \\
Covertype (min-max) & 256 & 10 & 0.975 & 0.971 & \textbf{0.972} & 0.972 & $-$0.3\% & 7/0/3 & 1.000 & $-$0.0\% & 4/2/4 & 1.000 \\
Covertype (min-max) & 1024 & 10 & 0.999 & 0.998 & \textbf{0.989} & 0.992 & $-$0.6\% & 9/0/1 & 0.285 & $-$0.2\% & 9/0/1 & 0.977 \\
Covertype (min-max) & 4096 & 10 & 0.997 & 1.002 & \textbf{0.998} & 0.999 & +0.0\% & 5/0/5 & 1.000 & $-$0.3\% & 9/1/0 & 0.285 \\
\bottomrule
\end{tabular}
}
\end{table}
\begin{table}[H]\centering\small
\caption{Random initialization, 5 epochs.}\label{tab:random-T5}\resizebox{\linewidth}{!}{\begin{tabular}{lllcccccccccc}
\toprule
 & & & \multicolumn{4}{c}{Median test MSE / median test MSE of Lloyd@25} & \multicolumn{3}{c}{\textsc{distance} vs.\ \textsc{count}} & \multicolumn{3}{c}{\textsc{distance} vs.\ \textsc{off}}\\
\cmidrule(lr){4-7}\cmidrule(lr){8-10}\cmidrule(lr){11-13}
Dataset & $k$ & Seeds & \textsc{count} & \textsc{off} & \textsc{distance} & \textsc{estimate} & $\Delta$ & W/T/L & $p$ & $\Delta$ & W/T/L & $p$\\
\midrule
\multicolumn{13}{l}{\emph{Held-out datasets}}\\
KDD Cup 99 & 256 & 10 & 1.027 & 0.616 & \textbf{0.583} & 0.579 & $-$41.2\% & 10/0/0 & 0.504 & $-$2.9\% & 7/0/3 & 1.000 \\
KDD Cup 99 & 1024 & 10 & 2.068 & 0.580 & \textbf{0.419} & 0.421 & $-$78.5\% & 10/0/0 & 0.504 & $-$17.0\% & 10/0/0 & 0.504 \\
KDD Cup 99 & 4096 & 10 & 1.092 & 0.779 & \textbf{0.690} & 0.713 & $-$36.1\% & 10/0/0 & 0.504 & $-$9.6\% & 9/0/1 & 0.508 \\
Census 1990 & 256 & 10 & 0.992 & 0.994 & \textbf{0.991} & 0.992 & +0.0\% & 1/9/0 & 1.000 & $-$0.3\% & 9/1/0 & 0.504 \\
Census 1990 & 1024 & 10 & 0.994 & 0.995 & \textbf{0.994} & 0.994 & +0.0\% & 3/3/4 & 1.000 & $-$0.0\% & 6/1/3 & 1.000 \\
Census 1990 & 4096 & 10 & 0.988 & 0.987 & \textbf{0.987} & 0.988 & $-$0.2\% & 7/2/1 & 1.000 & +0.0\% & 4/0/6 & 1.000 \\
YearPredictionMSD & 256 & 10 & 1.011 & 1.013 & \textbf{1.009} & 1.006 & $-$0.2\% & 6/1/3 & 1.000 & $-$0.3\% & 9/1/0 & 0.508 \\
YearPredictionMSD & 1024 & 10 & 1.006 & 1.032 & \textbf{1.021} & 1.004 & +1.7\% & 0/0/10 & 0.504 & $-$1.1\% & 10/0/0 & 0.504 \\
YearPredictionMSD & 4096 & 10 & 0.998 & 1.025 & \textbf{1.004} & 0.998 & +0.6\% & 0/0/10 & 0.504 & $-$2.2\% & 10/0/0 & 0.504 \\
GeoNames (2-D) & 256 & 10 & 1.008 & 0.906 & \textbf{0.890} & 0.916 & $-$10.1\% & 10/0/0 & 0.504 & +0.0\% & 4/4/2 & 1.000 \\
GeoNames (2-D) & 1024 & 10 & 1.185 & 0.856 & \textbf{0.868} & 0.866 & $-$26.8\% & 10/0/0 & 0.504 & $-$0.0\% & 6/0/4 & 1.000 \\
GeoNames (2-D) & 4096 & 10 & 1.047 & 0.906 & \textbf{0.909} & 0.893 & $-$13.6\% & 10/0/0 & 0.504 & $-$0.2\% & 6/0/4 & 1.000 \\
Household power & 256 & 10 & 1.000 & 0.933 & \textbf{0.936} & 0.931 & $-$6.5\% & 10/0/0 & 0.504 & +0.2\% & 4/1/5 & 1.000 \\
Household power & 1024 & 10 & 0.935 & 0.923 & \textbf{0.922} & 0.921 & $-$0.6\% & 6/0/4 & 1.000 & +0.4\% & 4/0/6 & 1.000 \\
Shuttle & 256 & 10 & 3.205 & 0.930 & \textbf{0.969} & 0.970 & $-$69.1\% & 10/0/0 & 0.504 & +2.7\% & 1/0/9 & 1.000 \\
Shuttle & 1024 & 10 & 1.584 & 1.005 & \textbf{1.019} & 1.021 & $-$36.3\% & 10/0/0 & 0.504 & +0.7\% & 3/0/7 & 1.000 \\
\midrule
\multicolumn{13}{l}{\emph{Development datasets}}\\
Covertype (z-score) & 256 & 10 & 1.267 & 0.763 & \textbf{0.746} & 0.745 & $-$39.1\% & 10/0/0 & 0.223 & $-$0.9\% & 10/0/0 & 0.223 \\
Covertype (z-score) & 1024 & 10 & 0.981 & 0.988 & \textbf{0.976} & 0.977 & $-$0.1\% & 8/0/2 & 1.000 & $-$0.8\% & 10/0/0 & 0.223 \\
Covertype (z-score) & 4096 & 10 & 0.994 & 0.994 & \textbf{0.993} & 0.993 & $-$0.1\% & 8/1/1 & 0.977 & $-$0.1\% & 5/2/3 & 1.000 \\
Covertype (min-max) & 256 & 10 & 0.989 & 0.982 & \textbf{0.978} & 0.978 & $-$0.8\% & 9/0/1 & 0.977 & $-$0.2\% & 8/1/1 & 1.000 \\
Covertype (min-max) & 1024 & 10 & 0.995 & 0.995 & \textbf{0.986} & 0.990 & $-$0.8\% & 8/0/2 & 0.625 & $-$0.3\% & 9/0/1 & 0.793 \\
Covertype (min-max) & 4096 & 10 & 0.989 & 0.993 & \textbf{0.990} & 0.989 & +0.0\% & 4/1/5 & 1.000 & $-$0.3\% & 10/0/0 & 0.223 \\
\bottomrule
\end{tabular}
}
\end{table}
\begin{table}[H]\centering\small
\caption{Random initialization, 10 epochs.}\label{tab:random-T10}\resizebox{\linewidth}{!}{\begin{tabular}{lllcccccccccc}
\toprule
 & & & \multicolumn{4}{c}{Median test MSE / median test MSE of Lloyd@25} & \multicolumn{3}{c}{\textsc{distance} vs.\ \textsc{count}} & \multicolumn{3}{c}{\textsc{distance} vs.\ \textsc{off}}\\
\cmidrule(lr){4-7}\cmidrule(lr){8-10}\cmidrule(lr){11-13}
Dataset & $k$ & Seeds & \textsc{count} & \textsc{off} & \textsc{distance} & \textsc{estimate} & $\Delta$ & W/T/L & $p$ & $\Delta$ & W/T/L & $p$\\
\midrule
\multicolumn{13}{l}{\emph{Held-out datasets}}\\
KDD Cup 99 & 256 & 10 & 0.938 & 0.578 & \textbf{0.521} & 0.547 & $-$41.9\% & 10/0/0 & 0.504 & $-$4.3\% & 7/0/3 & 1.000 \\
KDD Cup 99 & 1024 & 10 & 1.428 & 0.585 & \textbf{0.435} & 0.443 & $-$59.6\% & 10/0/0 & 0.504 & $-$15.4\% & 10/0/0 & 0.504 \\
KDD Cup 99 & 4096 & 10 & 1.145 & 0.755 & \textbf{0.670} & 0.681 & $-$38.9\% & 10/0/0 & 0.504 & $-$9.9\% & 10/0/0 & 0.504 \\
Census 1990 & 256 & 10 & 0.994 & 0.996 & \textbf{0.994} & 0.994 & +0.0\% & 0/10/0 & 1.000 & $-$0.2\% & 8/1/1 & 1.000 \\
Census 1990 & 1024 & 10 & 0.997 & 0.998 & \textbf{0.997} & 0.997 & +0.0\% & 3/5/2 & 1.000 & $-$0.0\% & 4/2/4 & 1.000 \\
Census 1990 & 4096 & 10 & 0.991 & 0.991 & \textbf{0.991} & 0.991 & $-$0.0\% & 5/0/5 & 1.000 & +0.0\% & 2/2/6 & 1.000 \\
YearPredictionMSD & 256 & 10 & 1.013 & 1.003 & \textbf{1.004} & 1.003 & $-$0.9\% & 10/0/0 & 0.504 & +0.1\% & 4/0/6 & 1.000 \\
YearPredictionMSD & 1024 & 10 & 1.002 & 1.018 & \textbf{1.013} & 1.002 & +1.1\% & 0/0/10 & 0.504 & $-$0.4\% & 10/0/0 & 0.504 \\
YearPredictionMSD & 4096 & 10 & 0.997 & 1.028 & \textbf{1.010} & 0.998 & +1.5\% & 0/0/10 & 0.504 & $-$1.7\% & 10/0/0 & 0.504 \\
GeoNames (2-D) & 256 & 10 & 1.007 & 0.896 & \textbf{0.887} & 0.918 & $-$11.3\% & 10/0/0 & 0.504 & $-$0.4\% & 6/3/1 & 1.000 \\
GeoNames (2-D) & 1024 & 10 & 1.170 & 0.846 & \textbf{0.856} & 0.859 & $-$24.7\% & 10/0/0 & 0.504 & $-$0.2\% & 5/0/5 & 1.000 \\
GeoNames (2-D) & 4096 & 10 & 1.020 & 0.896 & \textbf{0.886} & 0.891 & $-$15.4\% & 10/0/0 & 0.504 & $-$0.6\% & 7/0/3 & 1.000 \\
Household power & 256 & 10 & 0.996 & 0.929 & \textbf{0.935} & 0.929 & $-$4.5\% & 10/0/0 & 0.504 & +0.2\% & 1/1/8 & 1.000 \\
Household power & 1024 & 10 & 0.916 & 0.891 & \textbf{0.902} & 0.902 & $-$1.3\% & 8/1/1 & 1.000 & +0.2\% & 5/0/5 & 1.000 \\
Shuttle & 256 & 10 & 3.587 & 0.917 & \textbf{0.926} & 0.937 & $-$74.0\% & 10/0/0 & 0.504 & +1.0\% & 1/0/9 & 0.508 \\
Shuttle & 1024 & 10 & 1.253 & 0.948 & \textbf{0.952} & 0.949 & $-$24.0\% & 10/0/0 & 0.504 & +0.2\% & 1/1/8 & 0.721 \\
\midrule
\multicolumn{13}{l}{\emph{Development datasets}}\\
Covertype (z-score) & 256 & 10 & 1.166 & 1.009 & \textbf{0.998} & 0.998 & $-$20.3\% & 8/0/2 & 0.793 & $-$0.7\% & 10/0/0 & 0.223 \\
Covertype (z-score) & 1024 & 10 & 0.982 & 0.988 & \textbf{0.977} & 0.976 & $-$0.6\% & 10/0/0 & 0.223 & $-$0.9\% & 10/0/0 & 0.223 \\
Covertype (z-score) & 4096 & 10 & 0.987 & 0.991 & \textbf{0.987} & 0.986 & $-$0.0\% & 5/1/4 & 1.000 & $-$0.2\% & 8/0/2 & 0.625 \\
Covertype (min-max) & 256 & 10 & 1.031 & 1.021 & \textbf{1.020} & 1.023 & $-$0.6\% & 9/0/1 & 0.285 & $-$0.1\% & 6/4/0 & 0.819 \\
Covertype (min-max) & 1024 & 10 & 0.990 & 0.984 & \textbf{0.981} & 0.983 & $-$0.6\% & 8/0/2 & 0.625 & $-$0.2\% & 9/0/1 & 1.000 \\
Covertype (min-max) & 4096 & 10 & 0.986 & 0.988 & \textbf{0.983} & 0.984 & $-$0.0\% & 5/0/5 & 1.000 & $-$0.3\% & 10/0/0 & 0.223 \\
\bottomrule
\end{tabular}
}
\end{table}
\begin{table}[H]\centering\small
\caption{Smallest tested number of epochs at which a method's median test MSE is below the median test MSE of Lloyd@25, from random initialization. Each number of epochs is a separately trained run; ``$>10$'': not reached with any tested number of epochs up to 10.}\label{tab:epochs-random}\begin{tabular}{llcccc}
\toprule
Dataset & $k$ & \textsc{count} & \textsc{off} & \textsc{distance} & \textsc{estimate}\\
\midrule
KDD Cup 99 & 256 & 10 & 1 & 1 & 1\\
KDD Cup 99 & 1024 & $>10$ & 2 & 1 & 2\\
KDD Cup 99 & 4096 & $>10$ & 2 & 2 & 2\\
Census 1990 & 256 & 1 & 1 & 1 & 1\\
Census 1990 & 1024 & 2 & 1 & 1 & 2\\
Census 1990 & 4096 & 2 & 2 & 2 & 2\\
YearPredictionMSD & 256 & $>10$ & $>10$ & $>10$ & $>10$\\
YearPredictionMSD & 1024 & $>10$ & $>10$ & $>10$ & $>10$\\
YearPredictionMSD & 4096 & 5 & $>10$ & $>10$ & 5\\
GeoNames (2-D) & 256 & $>10$ & 2 & 2 & 2\\
GeoNames (2-D) & 1024 & $>10$ & 2 & 2 & 2\\
GeoNames (2-D) & 4096 & $>10$ & 2 & 2 & 2\\
Household power & 256 & 5 & 1 & 1 & 1\\
Household power & 1024 & 2 & 2 & 2 & 2\\
Shuttle & 256 & $>10$ & 5 & 5 & 5\\
Shuttle & 1024 & $>10$ & 10 & 10 & 10\\
Covertype (z-score) & 256 & $>10$ & 1 & 1 & 1\\
Covertype (z-score) & 1024 & 3 & 3 & 2 & 2\\
Covertype (z-score) & 4096 & 3 & 5 & 5 & 5\\
Covertype (min-max) & 256 & 1 & 1 & 1 & 1\\
Covertype (min-max) & 1024 & 3 & 3 & 3 & 3\\
Covertype (min-max) & 4096 & 3 & 5 & 3 & 3\\
\bottomrule
\end{tabular}

\end{table}

\subsection{Non-Embedding Datasets, k-means++ Initialization}\label{app:kmpp}
\Crefrange{tab:kmpp-T1}{tab:kmpp-T10} repeat the comparison from k-means++ initialization, with 1, 2, 3, 5 and 10 epochs (\cref{sec:exp-kmpp}); Lloyd@25 starts from the same k-means++ centres as the e-means methods. \Cref{tab:epochs-kmpp} is the counterpart of \cref{tab:epochs-random}.

\begin{table}[H]\centering\small
\caption{k-means++ initialization, 1 epoch. Lloyd@25 starts from the same k-means++ centres.}\label{tab:kmpp-T1}\resizebox{\linewidth}{!}{\begin{tabular}{lllcccccccccc}
\toprule
 & & & \multicolumn{4}{c}{Median test MSE / median test MSE of Lloyd@25} & \multicolumn{3}{c}{\textsc{distance} vs.\ \textsc{count}} & \multicolumn{3}{c}{\textsc{distance} vs.\ \textsc{off}}\\
\cmidrule(lr){4-7}\cmidrule(lr){8-10}\cmidrule(lr){11-13}
Dataset & $k$ & Seeds & \textsc{count} & \textsc{off} & \textsc{distance} & \textsc{estimate} & $\Delta$ & W/T/L & $p$ & $\Delta$ & W/T/L & $p$\\
\midrule
\multicolumn{13}{l}{\emph{Held-out datasets}}\\
KDD Cup 99 & 256 & 10 & 14.582 & 1.026 & \textbf{1.027} & 1.027 & $-$93.0\% & 10/0/0 & 0.504 & +0.0\% & 0/9/1 & 1.000 \\
KDD Cup 99 & 1024 & 10 & 30.804 & 1.007 & \textbf{1.007} & 1.005 & $-$96.7\% & 10/0/0 & 0.504 & +0.0\% & 3/2/5 & 1.000 \\
KDD Cup 99 & 4096 & 10 & 9.445 & 1.013 & \textbf{1.013} & 1.020 & $-$88.5\% & 10/0/0 & 0.504 & +0.0\% & 3/1/6 & 1.000 \\
Census 1990 & 256 & 10 & 1.001 & 1.001 & \textbf{1.001} & 1.001 & +0.0\% & 3/5/2 & 1.000 & +0.1\% & 3/0/7 & 1.000 \\
Census 1990 & 1024 & 10 & 1.004 & 1.000 & \textbf{1.002} & 1.003 & $-$0.2\% & 10/0/0 & 0.504 & +0.3\% & 1/0/9 & 1.000 \\
Census 1990 & 4096 & 10 & 1.013 & 1.006 & \textbf{1.011} & 1.012 & $-$0.2\% & 10/0/0 & 0.504 & +0.3\% & 0/0/10 & 0.504 \\
YearPredictionMSD & 256 & 10 & 1.039 & 1.017 & \textbf{1.017} & 1.023 & $-$2.0\% & 10/0/0 & 0.504 & +0.1\% & 2/0/8 & 1.000 \\
YearPredictionMSD & 1024 & 10 & 1.024 & 1.022 & \textbf{1.018} & 1.019 & $-$0.6\% & 10/0/0 & 0.504 & $-$0.4\% & 10/0/0 & 0.504 \\
YearPredictionMSD & 4096 & 10 & 1.010 & 1.032 & \textbf{1.017} & 1.017 & +0.8\% & 0/0/10 & 0.504 & $-$1.4\% & 10/0/0 & 0.504 \\
GeoNames (2-D) & 256 & 10 & 1.389 & 0.999 & \textbf{0.999} & 1.003 & $-$27.8\% & 10/0/0 & 0.504 & +0.0\% & 1/8/1 & 1.000 \\
GeoNames (2-D) & 1024 & 10 & 1.761 & 1.007 & \textbf{1.011} & 1.020 & $-$42.6\% & 10/0/0 & 0.504 & +0.2\% & 4/0/6 & 1.000 \\
GeoNames (2-D) & 4096 & 10 & 1.918 & 1.024 & \textbf{1.040} & 1.054 & $-$45.8\% & 10/0/0 & 0.504 & +1.3\% & 0/0/10 & 0.504 \\
\midrule
\multicolumn{13}{l}{\emph{Development datasets}}\\
Covertype (z-score) & 256 & 10 & 2.589 & 1.003 & \textbf{1.003} & 1.003 & $-$61.3\% & 10/0/0 & 0.223 & +0.0\% & 3/4/3 & 1.000 \\
Covertype (z-score) & 1024 & 10 & 1.069 & 1.011 & \textbf{1.011} & 1.011 & $-$5.2\% & 10/0/0 & 0.223 & +0.0\% & 4/0/6 & 1.000 \\
Covertype (z-score) & 4096 & 10 & 1.033 & 1.025 & \textbf{1.027} & 1.028 & $-$0.6\% & 10/0/0 & 0.223 & +0.2\% & 1/1/8 & 0.625 \\
Covertype (min-max) & 256 & 10 & 1.050 & 1.004 & \textbf{1.004} & 1.004 & $-$4.5\% & 10/0/0 & 0.223 & +0.0\% & 4/3/3 & 1.000 \\
Covertype (min-max) & 1024 & 10 & 1.034 & 1.011 & \textbf{1.011} & 1.010 & $-$2.5\% & 10/0/0 & 0.223 & $-$0.0\% & 5/2/3 & 1.000 \\
Covertype (min-max) & 4096 & 10 & 1.030 & 1.021 & \textbf{1.022} & 1.024 & $-$0.9\% & 10/0/0 & 0.223 & +0.2\% & 2/1/7 & 1.000 \\
\bottomrule
\end{tabular}
}
\end{table}
\begin{table}[H]\centering\small
\caption{k-means++ initialization, 2 epochs.}\label{tab:kmpp-T2}\resizebox{\linewidth}{!}{\begin{tabular}{lllcccccccccc}
\toprule
 & & & \multicolumn{4}{c}{Median test MSE / median test MSE of Lloyd@25} & \multicolumn{3}{c}{\textsc{distance} vs.\ \textsc{count}} & \multicolumn{3}{c}{\textsc{distance} vs.\ \textsc{off}}\\
\cmidrule(lr){4-7}\cmidrule(lr){8-10}\cmidrule(lr){11-13}
Dataset & $k$ & Seeds & \textsc{count} & \textsc{off} & \textsc{distance} & \textsc{estimate} & $\Delta$ & W/T/L & $p$ & $\Delta$ & W/T/L & $p$\\
\midrule
\multicolumn{13}{l}{\emph{Held-out datasets}}\\
KDD Cup 99 & 256 & 10 & 13.073 & 1.016 & \textbf{1.016} & 1.014 & $-$92.2\% & 10/0/0 & 0.504 & +0.0\% & 1/9/0 & 1.000 \\
KDD Cup 99 & 1024 & 10 & 25.672 & 1.006 & \textbf{1.006} & 1.005 & $-$96.1\% & 10/0/0 & 0.504 & $-$0.0\% & 6/1/3 & 1.000 \\
KDD Cup 99 & 4096 & 10 & 4.466 & 1.008 & \textbf{1.007} & 1.007 & $-$77.6\% & 10/0/0 & 0.504 & $-$0.0\% & 6/1/3 & 1.000 \\
Census 1990 & 256 & 10 & 0.996 & 0.996 & \textbf{0.996} & 0.996 & +0.0\% & 4/5/1 & 1.000 & +0.0\% & 3/1/6 & 1.000 \\
Census 1990 & 1024 & 10 & 0.997 & 0.995 & \textbf{0.996} & 0.997 & $-$0.1\% & 10/0/0 & 0.504 & +0.1\% & 2/0/8 & 1.000 \\
Census 1990 & 4096 & 10 & 1.002 & 0.997 & \textbf{0.999} & 1.001 & $-$0.2\% & 10/0/0 & 0.504 & +0.2\% & 0/0/10 & 0.504 \\
YearPredictionMSD & 256 & 10 & 1.020 & 1.029 & \textbf{1.025} & 1.011 & +0.4\% & 1/0/9 & 1.000 & $-$0.5\% & 10/0/0 & 0.504 \\
YearPredictionMSD & 1024 & 10 & 1.008 & 1.025 & \textbf{1.017} & 1.007 & +1.0\% & 0/0/10 & 0.504 & $-$0.8\% & 10/0/0 & 0.504 \\
YearPredictionMSD & 4096 & 10 & 0.997 & 1.026 & \textbf{1.012} & 1.003 & +1.4\% & 0/0/10 & 0.504 & $-$1.4\% & 10/0/0 & 0.504 \\
GeoNames (2-D) & 256 & 10 & 1.304 & 1.003 & \textbf{1.003} & 1.003 & $-$23.2\% & 10/0/0 & 0.504 & +0.0\% & 0/10/0 & 1.000 \\
GeoNames (2-D) & 1024 & 10 & 1.755 & 0.994 & \textbf{0.996} & 1.001 & $-$43.5\% & 10/0/0 & 0.504 & +0.0\% & 2/6/2 & 1.000 \\
GeoNames (2-D) & 4096 & 10 & 1.603 & 1.007 & \textbf{1.008} & 1.017 & $-$37.9\% & 10/0/0 & 0.504 & +0.1\% & 3/0/7 & 1.000 \\
\midrule
\multicolumn{13}{l}{\emph{Development datasets}}\\
Covertype (z-score) & 256 & 10 & 2.262 & 0.998 & \textbf{0.998} & 0.998 & $-$55.8\% & 10/0/0 & 0.223 & +0.0\% & 4/4/2 & 1.000 \\
Covertype (z-score) & 1024 & 10 & 1.006 & 0.998 & \textbf{0.999} & 0.998 & $-$0.7\% & 10/0/0 & 0.223 & $-$0.0\% & 6/1/3 & 1.000 \\
Covertype (z-score) & 4096 & 10 & 1.009 & 1.007 & \textbf{1.007} & 1.007 & $-$0.2\% & 10/0/0 & 0.223 & +0.1\% & 2/0/8 & 1.000 \\
Covertype (min-max) & 256 & 10 & 1.044 & 0.998 & \textbf{0.996} & 0.996 & $-$3.8\% & 10/0/0 & 0.223 & $-$0.1\% & 8/1/1 & 0.691 \\
Covertype (min-max) & 1024 & 10 & 1.018 & 1.000 & \textbf{1.001} & 1.000 & $-$1.7\% & 10/0/0 & 0.223 & $-$0.0\% & 5/1/4 & 1.000 \\
Covertype (min-max) & 4096 & 10 & 1.010 & 1.004 & \textbf{1.004} & 1.004 & $-$0.5\% & 10/0/0 & 0.223 & +0.0\% & 4/0/6 & 1.000 \\
\bottomrule
\end{tabular}
}
\end{table}
\begin{table}[H]\centering\small
\caption{k-means++ initialization, 3 epochs.}\label{tab:kmpp-T3}\resizebox{\linewidth}{!}{\begin{tabular}{lllcccccccccc}
\toprule
 & & & \multicolumn{4}{c}{Median test MSE / median test MSE of Lloyd@25} & \multicolumn{3}{c}{\textsc{distance} vs.\ \textsc{count}} & \multicolumn{3}{c}{\textsc{distance} vs.\ \textsc{off}}\\
\cmidrule(lr){4-7}\cmidrule(lr){8-10}\cmidrule(lr){11-13}
Dataset & $k$ & Seeds & \textsc{count} & \textsc{off} & \textsc{distance} & \textsc{estimate} & $\Delta$ & W/T/L & $p$ & $\Delta$ & W/T/L & $p$\\
\midrule
\multicolumn{13}{l}{\emph{Held-out datasets}}\\
KDD Cup 99 & 256 & 10 & 12.796 & 1.009 & \textbf{1.009} & 1.013 & $-$92.1\% & 10/0/0 & 0.504 & +0.0\% & 1/9/0 & 1.000 \\
KDD Cup 99 & 1024 & 10 & 23.796 & 1.004 & \textbf{1.003} & 1.002 & $-$95.8\% & 10/0/0 & 0.504 & $-$0.1\% & 8/0/2 & 1.000 \\
KDD Cup 99 & 4096 & 10 & 3.884 & 1.006 & \textbf{1.006} & 1.009 & $-$74.1\% & 10/0/0 & 0.504 & $-$0.0\% & 6/0/4 & 1.000 \\
Census 1990 & 256 & 10 & 0.998 & 0.998 & \textbf{0.998} & 0.998 & +0.0\% & 0/9/1 & 1.000 & +0.0\% & 4/3/3 & 1.000 \\
Census 1990 & 1024 & 10 & 0.997 & 0.994 & \textbf{0.995} & 0.996 & $-$0.2\% & 10/0/0 & 0.504 & +0.1\% & 1/0/9 & 1.000 \\
Census 1990 & 4096 & 10 & 1.000 & 0.996 & \textbf{0.997} & 1.000 & $-$0.3\% & 10/0/0 & 0.504 & +0.2\% & 1/1/8 & 0.721 \\
YearPredictionMSD & 256 & 10 & 1.016 & 1.035 & \textbf{1.028} & 1.009 & +1.0\% & 0/0/10 & 0.504 & $-$0.7\% & 10/0/0 & 0.504 \\
YearPredictionMSD & 1024 & 10 & 1.006 & 1.029 & \textbf{1.018} & 1.005 & +1.2\% & 0/0/10 & 0.504 & $-$1.1\% & 10/0/0 & 0.504 \\
YearPredictionMSD & 4096 & 10 & 0.995 & 1.022 & \textbf{1.011} & 1.001 & +1.7\% & 0/0/10 & 0.504 & $-$1.1\% & 10/0/0 & 0.504 \\
GeoNames (2-D) & 256 & 10 & 1.301 & 0.995 & \textbf{0.995} & 0.996 & $-$23.9\% & 10/0/0 & 0.504 & +0.0\% & 0/10/0 & 1.000 \\
GeoNames (2-D) & 1024 & 10 & 1.694 & 0.993 & \textbf{0.993} & 0.995 & $-$41.1\% & 10/0/0 & 0.504 & $-$0.0\% & 4/3/3 & 1.000 \\
GeoNames (2-D) & 4096 & 10 & 1.565 & 1.005 & \textbf{1.005} & 1.014 & $-$35.9\% & 10/0/0 & 0.504 & +0.1\% & 4/0/6 & 1.000 \\
\midrule
\multicolumn{13}{l}{\emph{Development datasets}}\\
Covertype (z-score) & 256 & 10 & 2.212 & 0.998 & \textbf{0.997} & 0.997 & $-$54.6\% & 10/0/0 & 0.223 & $-$0.0\% & 6/3/1 & 1.000 \\
Covertype (z-score) & 1024 & 10 & 1.006 & 0.997 & \textbf{0.997} & 0.997 & $-$1.0\% & 10/0/0 & 0.223 & $-$0.0\% & 5/0/5 & 1.000 \\
Covertype (z-score) & 4096 & 10 & 1.001 & 1.000 & \textbf{0.999} & 0.999 & $-$0.2\% & 8/0/2 & 0.793 & $-$0.0\% & 5/0/5 & 1.000 \\
Covertype (min-max) & 256 & 10 & 1.043 & 0.996 & \textbf{0.994} & 0.994 & $-$3.9\% & 10/0/0 & 0.223 & $-$0.1\% & 8/1/1 & 1.000 \\
Covertype (min-max) & 1024 & 10 & 1.016 & 0.999 & \textbf{0.998} & 0.998 & $-$1.8\% & 10/0/0 & 0.223 & $-$0.0\% & 6/2/2 & 1.000 \\
Covertype (min-max) & 4096 & 10 & 1.004 & 1.001 & \textbf{0.999} & 1.000 & $-$0.5\% & 10/0/0 & 0.223 & $-$0.1\% & 8/0/2 & 1.000 \\
\bottomrule
\end{tabular}
}
\end{table}
\begin{table}[H]\centering\small
\caption{k-means++ initialization, 5 epochs.}\label{tab:kmpp-T5}\resizebox{\linewidth}{!}{\begin{tabular}{lllcccccccccc}
\toprule
 & & & \multicolumn{4}{c}{Median test MSE / median test MSE of Lloyd@25} & \multicolumn{3}{c}{\textsc{distance} vs.\ \textsc{count}} & \multicolumn{3}{c}{\textsc{distance} vs.\ \textsc{off}}\\
\cmidrule(lr){4-7}\cmidrule(lr){8-10}\cmidrule(lr){11-13}
Dataset & $k$ & Seeds & \textsc{count} & \textsc{off} & \textsc{distance} & \textsc{estimate} & $\Delta$ & W/T/L & $p$ & $\Delta$ & W/T/L & $p$\\
\midrule
\multicolumn{13}{l}{\emph{Held-out datasets}}\\
KDD Cup 99 & 256 & 10 & 11.748 & 1.014 & \textbf{1.014} & 1.014 & $-$91.2\% & 10/0/0 & 0.504 & +0.0\% & 0/10/0 & 1.000 \\
KDD Cup 99 & 1024 & 10 & 19.774 & 1.004 & \textbf{1.003} & 1.003 & $-$94.9\% & 10/0/0 & 0.504 & $-$0.1\% & 6/3/1 & 1.000 \\
KDD Cup 99 & 4096 & 10 & 4.204 & 1.000 & \textbf{1.000} & 1.005 & $-$76.4\% & 10/0/0 & 0.504 & $-$0.0\% & 7/1/2 & 1.000 \\
Census 1990 & 256 & 10 & 0.998 & 0.998 & \textbf{0.998} & 0.998 & +0.0\% & 0/9/1 & 1.000 & +0.0\% & 4/1/5 & 1.000 \\
Census 1990 & 1024 & 10 & 0.996 & 0.996 & \textbf{0.996} & 0.996 & $-$0.1\% & 8/0/2 & 1.000 & $-$0.0\% & 6/0/4 & 1.000 \\
Census 1990 & 4096 & 10 & 0.999 & 0.995 & \textbf{0.995} & 0.998 & $-$0.3\% & 10/0/0 & 0.504 & +0.1\% & 2/1/7 & 1.000 \\
YearPredictionMSD & 256 & 10 & 1.013 & 1.016 & \textbf{1.013} & 1.005 & $-$0.0\% & 6/0/4 & 1.000 & $-$0.3\% & 10/0/0 & 0.504 \\
YearPredictionMSD & 1024 & 10 & 1.005 & 1.036 & \textbf{1.027} & 1.004 & +2.1\% & 0/0/10 & 0.504 & $-$0.9\% & 10/0/0 & 0.504 \\
YearPredictionMSD & 4096 & 10 & 0.994 & 1.025 & \textbf{1.012} & 0.999 & +1.8\% & 0/0/10 & 0.504 & $-$1.4\% & 10/0/0 & 0.504 \\
GeoNames (2-D) & 256 & 10 & 1.310 & 0.998 & \textbf{0.998} & 0.998 & $-$24.2\% & 10/0/0 & 0.504 & +0.0\% & 0/10/0 & 1.000 \\
GeoNames (2-D) & 1024 & 10 & 1.575 & 0.991 & \textbf{0.992} & 0.992 & $-$37.2\% & 10/0/0 & 0.504 & +0.0\% & 2/4/4 & 1.000 \\
GeoNames (2-D) & 4096 & 10 & 1.625 & 0.999 & \textbf{1.000} & 1.007 & $-$38.2\% & 10/0/0 & 0.504 & +0.2\% & 0/0/10 & 0.504 \\
\midrule
\multicolumn{13}{l}{\emph{Development datasets}}\\
Covertype (z-score) & 256 & 10 & 1.697 & 0.998 & \textbf{0.998} & 0.998 & $-$41.2\% & 10/0/0 & 0.223 & +0.0\% & 3/6/1 & 1.000 \\
Covertype (z-score) & 1024 & 10 & 1.002 & 0.993 & \textbf{0.992} & 0.992 & $-$1.0\% & 10/0/0 & 0.223 & $-$0.1\% & 6/0/4 & 1.000 \\
Covertype (z-score) & 4096 & 10 & 0.997 & 0.994 & \textbf{0.994} & 0.995 & $-$0.2\% & 6/0/4 & 1.000 & +0.0\% & 2/1/7 & 1.000 \\
Covertype (min-max) & 256 & 10 & 1.033 & 0.997 & \textbf{0.995} & 0.995 & $-$3.7\% & 10/0/0 & 0.223 & $-$0.1\% & 7/3/0 & 0.819 \\
Covertype (min-max) & 1024 & 10 & 1.015 & 0.995 & \textbf{0.995} & 0.994 & $-$1.9\% & 10/0/0 & 0.223 & $-$0.0\% & 6/1/3 & 1.000 \\
Covertype (min-max) & 4096 & 10 & 1.000 & 0.995 & \textbf{0.994} & 0.995 & $-$0.5\% & 10/0/0 & 0.223 & $-$0.1\% & 8/0/2 & 1.000 \\
\bottomrule
\end{tabular}
}
\end{table}
\begin{table}[H]\centering\small
\caption{k-means++ initialization, 10 epochs.}\label{tab:kmpp-T10}\resizebox{\linewidth}{!}{\begin{tabular}{lllcccccccccc}
\toprule
 & & & \multicolumn{4}{c}{Median test MSE / median test MSE of Lloyd@25} & \multicolumn{3}{c}{\textsc{distance} vs.\ \textsc{count}} & \multicolumn{3}{c}{\textsc{distance} vs.\ \textsc{off}}\\
\cmidrule(lr){4-7}\cmidrule(lr){8-10}\cmidrule(lr){11-13}
Dataset & $k$ & Seeds & \textsc{count} & \textsc{off} & \textsc{distance} & \textsc{estimate} & $\Delta$ & W/T/L & $p$ & $\Delta$ & W/T/L & $p$\\
\midrule
\multicolumn{13}{l}{\emph{Held-out datasets}}\\
KDD Cup 99 & 256 & 10 & 11.516 & 1.019 & \textbf{1.019} & 1.017 & $-$91.2\% & 10/0/0 & 0.504 & +0.0\% & 0/10/0 & 1.000 \\
KDD Cup 99 & 1024 & 10 & 6.495 & 0.997 & \textbf{0.996} & 0.996 & $-$84.0\% & 10/0/0 & 0.504 & $-$0.1\% & 7/2/1 & 1.000 \\
KDD Cup 99 & 4096 & 10 & 4.605 & 1.002 & \textbf{1.004} & 1.010 & $-$78.4\% & 10/0/0 & 0.504 & +0.1\% & 3/1/6 & 1.000 \\
Census 1990 & 256 & 10 & 0.998 & 0.998 & \textbf{0.998} & 0.998 & +0.0\% & 0/8/2 & 1.000 & +0.0\% & 4/1/5 & 1.000 \\
Census 1990 & 1024 & 10 & 0.997 & 0.997 & \textbf{0.997} & 0.997 & $-$0.1\% & 7/1/2 & 1.000 & $-$0.1\% & 7/1/2 & 1.000 \\
Census 1990 & 4096 & 10 & 0.998 & 0.995 & \textbf{0.996} & 0.997 & $-$0.2\% & 10/0/0 & 0.504 & +0.1\% & 3/0/7 & 1.000 \\
YearPredictionMSD & 256 & 10 & 1.011 & 1.006 & \textbf{1.005} & 1.005 & $-$0.5\% & 10/0/0 & 0.504 & $-$0.1\% & 6/0/4 & 1.000 \\
YearPredictionMSD & 1024 & 10 & 1.001 & 1.021 & \textbf{1.016} & 1.001 & +1.6\% & 0/0/10 & 0.504 & $-$0.5\% & 10/0/0 & 0.504 \\
YearPredictionMSD & 4096 & 10 & 0.992 & 1.030 & \textbf{1.020} & 0.998 & +2.7\% & 0/0/10 & 0.504 & $-$1.1\% & 10/0/0 & 0.504 \\
GeoNames (2-D) & 256 & 10 & 1.315 & 0.994 & \textbf{0.994} & 0.992 & $-$24.3\% & 10/0/0 & 0.504 & +0.0\% & 0/10/0 & 1.000 \\
GeoNames (2-D) & 1024 & 10 & 1.743 & 0.985 & \textbf{0.985} & 0.984 & $-$43.2\% & 10/0/0 & 0.504 & +0.0\% & 1/8/1 & 1.000 \\
GeoNames (2-D) & 4096 & 10 & 1.600 & 0.998 & \textbf{0.997} & 1.000 & $-$37.8\% & 10/0/0 & 0.504 & +0.0\% & 5/0/5 & 1.000 \\
\midrule
\multicolumn{13}{l}{\emph{Development datasets}}\\
Covertype (z-score) & 256 & 10 & 1.833 & 0.998 & \textbf{0.998} & 0.998 & $-$45.7\% & 10/0/0 & 0.223 & +0.0\% & 2/7/1 & 1.000 \\
Covertype (z-score) & 1024 & 10 & 1.004 & 0.994 & \textbf{0.993} & 0.993 & $-$1.2\% & 10/0/0 & 0.223 & $-$0.1\% & 8/1/1 & 0.393 \\
Covertype (z-score) & 4096 & 10 & 0.994 & 0.991 & \textbf{0.992} & 0.992 & $-$0.3\% & 8/0/2 & 0.625 & +0.1\% & 3/0/7 & 1.000 \\
Covertype (min-max) & 256 & 10 & 1.035 & 1.002 & \textbf{1.001} & 1.001 & $-$3.6\% & 10/0/0 & 0.223 & $-$0.1\% & 5/4/1 & 1.000 \\
Covertype (min-max) & 1024 & 10 & 1.011 & 0.995 & \textbf{0.994} & 0.993 & $-$1.6\% & 10/0/0 & 0.223 & $-$0.1\% & 7/0/3 & 1.000 \\
Covertype (min-max) & 4096 & 10 & 0.997 & 0.991 & \textbf{0.991} & 0.991 & $-$0.6\% & 9/0/1 & 0.285 & $-$0.1\% & 6/1/3 & 1.000 \\
\bottomrule
\end{tabular}
}
\end{table}
\begin{table}[H]\centering\small
\caption{Smallest tested number of epochs at which a method's median test MSE is below the median test MSE of Lloyd@25, from k-means++ initialization (as in \cref{tab:epochs-random}).}\label{tab:epochs-kmpp}\begin{tabular}{llcccc}
\toprule
Dataset & $k$ & \textsc{count} & \textsc{off} & \textsc{distance} & \textsc{estimate}\\
\midrule
KDD Cup 99 & 256 & $>10$ & $>10$ & $>10$ & $>10$\\
KDD Cup 99 & 1024 & $>10$ & 10 & 10 & 10\\
KDD Cup 99 & 4096 & $>10$ & $>10$ & 5 & $>10$\\
Census 1990 & 256 & 2 & 2 & 2 & 2\\
Census 1990 & 1024 & 2 & 2 & 2 & 2\\
Census 1990 & 4096 & 5 & 2 & 2 & 3\\
YearPredictionMSD & 256 & $>10$ & $>10$ & $>10$ & $>10$\\
YearPredictionMSD & 1024 & $>10$ & $>10$ & $>10$ & $>10$\\
YearPredictionMSD & 4096 & 2 & $>10$ & $>10$ & 5\\
GeoNames (2-D) & 256 & $>10$ & 1 & 1 & 3\\
GeoNames (2-D) & 1024 & $>10$ & 2 & 2 & 3\\
GeoNames (2-D) & 4096 & $>10$ & 5 & 10 & 10\\
Covertype (z-score) & 256 & $>10$ & 2 & 2 & 2\\
Covertype (z-score) & 1024 & $>10$ & 2 & 2 & 2\\
Covertype (z-score) & 4096 & 5 & 3 & 3 & 3\\
Covertype (min-max) & 256 & $>10$ & 2 & 2 & 2\\
Covertype (min-max) & 1024 & $>10$ & 2 & 3 & 2\\
Covertype (min-max) & 4096 & 10 & 5 & 3 & 5\\
\bottomrule
\end{tabular}

\end{table}

\subsection{Bound versus Utility Estimate}\label{app:exact}
\Cref{tab:exact} compares \Dist{} (the bound) with \Est{} (the online utility estimate) on the tabular datasets and on DEEP~1M and SIFT~1M (\cref{sec:exp-exact}).

\begin{table}[H]\centering\small
\caption{\Dist{} (bound) versus \Est{} (utility estimate), random initialization: median over seeds of the relative change in the test MSE of \Dist{} with respect to \Est{} after 2 and 5 epochs, W/T/L after 2 epochs, and the median ratio of the running time of \Dist{} to that of \Est. These runs shared their machine with other jobs, so the times are noisy; \cref{sec:exp-cost} reports controlled timings.}\label{tab:exact}\begin{tabular}{llcccc}
\toprule
Dataset & $k$ & $\Delta$ test ($T=2$) & $\Delta$ test ($T=5$) & W/T/L ($T=2$) & Time ratio\\
\midrule
KDD Cup 99 & 256 & +0.1\% & +0.1\% & 13/0/17 & 0.72\\
KDD Cup 99 & 1024 & $-$0.1\% & $-$0.9\% & 16/0/14 & 0.53\\
KDD Cup 99 & 4096 & $-$1.6\% & $-$2.8\% & 20/0/10 & 0.48\\
Census 1990 & 256 & +0.0\% & +0.0\% & 3/19/8 & 0.82\\
Census 1990 & 1024 & $-$0.1\% & +0.0\% & 22/4/4 & 0.57\\
Census 1990 & 4096 & $-$0.1\% & $-$0.1\% & 26/0/4 & 0.71\\
YearPredictionMSD & 256 & +1.0\% & +0.4\% & 0/0/30 & 0.67\\
YearPredictionMSD & 1024 & +0.4\% & +1.7\% & 0/0/30 & 0.55\\
YearPredictionMSD & 4096 & +0.2\% & +0.6\% & 3/0/27 & 0.51\\
GeoNames (2-D) & 256 & +0.0\% & +0.0\% & 10/17/3 & 0.60\\
GeoNames (2-D) & 1024 & +0.0\% & +0.0\% & 14/3/13 & 0.55\\
GeoNames (2-D) & 4096 & $-$1.3\% & $-$0.2\% & 24/0/6 & 0.50\\
Household power & 256 & $-$0.2\% & +0.4\% & 6/1/3 & 0.61\\
Household power & 1024 & $-$0.5\% & +0.5\% & 8/0/2 & 0.56\\
Shuttle & 256 & $-$1.6\% & $-$0.6\% & 7/0/3 & 0.75\\
Shuttle & 1024 & +0.4\% & $-$0.0\% & 4/0/6 & 0.62\\
Covertype (z-score) & 256 & +0.1\% & +0.0\% & 11/1/18 & 0.72\\
Covertype (z-score) & 1024 & +0.1\% & +0.1\% & 10/0/20 & 0.62\\
Covertype (z-score) & 4096 & +0.0\% & $-$0.0\% & 14/1/15 & 0.61\\
Covertype (min-max) & 256 & +0.0\% & +0.0\% & 9/16/5 & 0.74\\
Covertype (min-max) & 1024 & +0.0\% & $-$0.0\% & 7/3/20 & 0.64\\
Covertype (min-max) & 4096 & +0.0\% & +0.1\% & 10/1/19 & 0.67\\
DEEP 1M & 16384 & $-$0.0\% & $-$0.0\% & 3/0/2 & 0.52\\
SIFT 1M & 16384 & +0.0\% & +0.0\% & 2/0/3 & 0.80\\
\bottomrule
\end{tabular}

\end{table}

\subsection{Sensitivity}\label{app:sens}
\Cref{tab:sens} gives the effect of the threshold $\taunear$, of the waiting period $W$ and of removing the abandonment clause (\cref{sec:exp-sens}).

\begin{table}[H]\centering\small
\caption{Sensitivity: median over seeds of the relative change (\%) in test MSE with respect to \Count, random initialization, seeds 0--9, $k=1024$ ($k=16{,}384$ for DEEP~1M). Columns: values of $\taunear$, waiting periods $W$ (in epochs), no abandonment clause, and \Off.}\label{tab:sens}\resizebox{\linewidth}{!}{\begin{tabular}{llcccccccccccc}
\toprule
Dataset & $T$ & $\taunear{=}1$ & 2 & 3 & \textbf{4} & 6 & 8 & 16 & 32 & $W{=}1$ & $W{=}3$ & No clause & \textsc{off}\\
\midrule
Covertype (z-score) & 2 & +0.0 & $-$0.2 & $-$0.2 & $-$0.3 & $-$0.3 & $-$0.3 & $-$0.3 & $-$0.1 & $-$0.3 & $-$0.3 & $-$0.3 & +0.6\\
Covertype (z-score) & 5 & $-$0.2 & $-$0.3 & $-$0.2 & $-$0.1 & $-$0.2 & $-$0.4 & $-$0.3 & $-$0.1 & $-$0.2 & $-$0.1 & $-$0.1 & +0.6\\
Covertype (z-score) & 10 & $-$0.4 & $-$0.5 & $-$0.6 & $-$0.6 & $-$0.6 & $-$0.5 & $-$0.5 & $-$0.2 & $-$0.5 & $-$0.4 & $-$0.3 & +0.5\\
KDD Cup 99 & 2 & $-$80.7 & $-$82.5 & $-$81.3 & $-$82.3 & $-$82.3 & $-$82.2 & $-$82.0 & $-$81.6 & $-$82.1 & $-$82.3 & $-$82.3 & $-$79.0\\
KDD Cup 99 & 5 & $-$79.3 & $-$79.6 & $-$78.9 & $-$78.5 & $-$78.3 & $-$78.3 & $-$78.3 & $-$78.3 & $-$76.8 & $-$78.5 & $-$78.5 & $-$72.6\\
KDD Cup 99 & 10 & $-$62.7 & $-$58.9 & $-$58.9 & $-$59.6 & $-$58.5 & $-$58.8 & $-$57.3 & $-$52.3 & $-$57.2 & $-$59.6 & $-$59.6 & $-$47.7\\
GeoNames (2-D) & 2 & $-$24.8 & $-$24.9 & $-$24.7 & $-$24.7 & $-$24.6 & $-$24.1 & $-$20.1 & $-$20.4 & $-$24.7 & $-$24.7 & $-$24.7 & $-$25.7\\
GeoNames (2-D) & 5 & $-$27.4 & $-$27.4 & $-$27.4 & $-$26.8 & $-$27.4 & $-$27.2 & $-$25.1 & $-$22.6 & $-$26.8 & $-$26.8 & $-$26.8 & $-$27.8\\
GeoNames (2-D) & 10 & $-$25.4 & $-$25.4 & $-$25.4 & $-$24.7 & $-$24.4 & $-$24.4 & $-$24.2 & $-$22.6 & $-$24.7 & $-$24.7 & $-$24.7 & $-$25.3\\
DEEP 1M & 2 & +2.1 & +0.0 & -- & +0.0 & -- & +0.0 & -- & -- & +0.0 & -- & -- & +3.3\\
DEEP 1M & 5 & +1.5 & +0.0 & -- & +0.0 & -- & +0.0 & -- & -- & +0.0 & -- & -- & +2.2\\
\bottomrule
\end{tabular}
}
\end{table}

\subsection{Cluster-Size Imbalance and Tail Error}\label{app:imbalance}
\Cref{tab:imbalance} gives two quantities other than the MSE that \cref{sec:exp-mech} discusses: how unequal the cluster sizes are, and the mean error of the points with the largest errors.

\begin{table}[H]\centering\small
\caption{Cluster-size imbalance $k\sum_j|P_j|^2/n^2$ (1 when all clusters have the same size) and mean of the largest 0.1\% of the per-point training errors; random initialization, 2 epochs, medians over seeds.}\label{tab:imbalance}\begin{tabular}{llcccc}
\toprule
 & & \multicolumn{2}{c}{Imbalance $k\sum_j |P_j|^2/n^2$} & \multicolumn{2}{c}{Mean of the worst 0.1\% errors} \\
\cmidrule(lr){3-4}\cmidrule(lr){5-6}
Dataset & $k$ & \textsc{count} & \textsc{distance} & \textsc{count} & \textsc{distance}\\
\midrule
KDD Cup 99 & 256 & 2.08 & 2.09 & 2.4e+04 & 1.09e+04\\
KDD Cup 99 & 1024 & 1.57 & 1.57 & 1.93e+04 & 4.16e+03\\
KDD Cup 99 & 4096 & 1.75 & 1.76 & 1.42e+04 & 2.48e+03\\
Census 1990 & 256 & 1.36 & 1.37 & 151 & 151\\
Census 1990 & 1024 & 1.29 & 1.30 & 129 & 127\\
Census 1990 & 4096 & 1.25 & 1.25 & 102 & 99.1\\
YearPredictionMSD & 256 & 1.50 & 2.20 & 2.46e+03 & 1.58e+03\\
YearPredictionMSD & 1024 & 1.43 & 1.78 & 2.02e+03 & 1.31e+03\\
YearPredictionMSD & 4096 & 1.38 & 1.56 & 1.58e+03 & 1.06e+03\\
GeoNames (2-D) & 256 & 1.24 & 1.27 & 2.04e+03 & 1.18e+03\\
GeoNames (2-D) & 1024 & 1.21 & 1.22 & 825 & 393\\
GeoNames (2-D) & 4096 & 1.17 & 1.17 & 235 & 167\\
Household power & 256 & 1.40 & 1.44 & 26.3 & 20\\
Household power & 1024 & 1.25 & 1.26 & 12.5 & 12.3\\
Shuttle & 256 & 1.27 & 1.28 & 1.67e+03 & 215\\
Shuttle & 1024 & 1.32 & 1.33 & 431 & 200\\
Covertype (z-score) & 256 & 1.37 & 1.35 & 3.71e+03 & 41.8\\
Covertype (z-score) & 1024 & 1.22 & 1.22 & 1.02e+03 & 27.7\\
Covertype (z-score) & 4096 & 1.17 & 1.18 & 965 & 12.8\\
Covertype (min-max) & 256 & 1.24 & 1.26 & 1.49 & 1.48\\
Covertype (min-max) & 1024 & 1.21 & 1.21 & 0.872 & 0.758\\
Covertype (min-max) & 4096 & 1.17 & 1.18 & 0.281 & 0.269\\
\bottomrule
\end{tabular}

\end{table}

\subsection{Synthetic Data}\label{app:synth}
We use two synthetic datasets, which \cref{sec:exp-mech} discusses. \Cref{tab:synth} gives the results after 5 epochs, over 10 seeds; errors are measured on the training points.

\paragraph{Far groups.} 100 Gaussian blobs of 2{,}000 points in $d=32$ dimensions and 5 far groups of $\nfar$ points each, centred at distance $20$ from the origin in random directions, with $k=105$. A far group counts as kept if a centre ends within distance 5 (a quarter of 20) of the group's centre.

\paragraph{Heterogeneous scales.} 90 tight blobs (standard deviation $0.1$) and 10 broad blobs (standard deviation $3$), with $k=100$, in $d=2$ and $d=16$ dimensions.

\begin{table}[H]\centering\small
\caption{Synthetic data, 5 epochs, 10 seeds; errors are measured on the training points. Far groups: mean number of far groups kept over seeds. Heterogeneous scales: median MSE.}\label{tab:synth}\resizebox{\linewidth}{!}{\begin{tabular}{llccccccc}
\toprule
 & & \multicolumn{5}{c}{Far groups kept (of 5), by group size $n_{\mathrm{far}}$} & \multicolumn{2}{c}{Heterogeneous, MSE} \\
\cmidrule(lr){3-7}\cmidrule(lr){8-9}
Initialization & Method & 50 & 100 & 150 & 250 & 400 & $d{=}2$ & $d{=}16$\\
\midrule
Random & Lloyd@25 & 2.3 & 2.7 & 2.8 & 3.3 & 4.7 & 0.822 & 180\\
Random & \textsc{count} & 0.0 & 0.0 & 0.0 & 0.7 & 5.0 & 0.562 & 152\\
Random & \textsc{off} & 2.8 & 2.5 & 3.0 & 3.4 & 5.0 & 0.622 & 198\\
Random & \textsc{distance} & 2.5 & 2.4 & 2.9 & 3.4 & 5.0 & 0.611 & 152\\
Random & \textsc{distance}, no abandonment & 2.5 & 2.4 & 2.9 & 3.4 & 5.0 & 0.611 & 152\\
Random & \textsc{estimate} & 2.4 & 2.4 & 2.9 & 3.4 & 5.0 & 0.571 & 158\\
\midrule
k-means++ & Lloyd@25 & 2.6 & 4.6 & 5.0 & 5.0 & 5.0 & 0.293 & 14.5\\
k-means++ & \textsc{count} & 0.0 & 0.0 & 0.0 & 0.6 & 4.9 & 0.333 & 14.5\\
k-means++ & \textsc{off} & 2.8 & 4.7 & 5.0 & 5.0 & 5.0 & 0.29 & 14.5\\
k-means++ & \textsc{distance} & 2.7 & 4.7 & 5.0 & 5.0 & 5.0 & 0.29 & 14.5\\
k-means++ & \textsc{distance}, no abandonment & 2.7 & 4.7 & 5.0 & 5.0 & 5.0 & 0.29 & 14.5\\
k-means++ & \textsc{estimate} & 2.6 & 4.7 & 5.0 & 5.0 & 5.0 & 0.288 & 14.5\\
\bottomrule
\end{tabular}
}
\end{table}

\section{Data and Protocol Details}\label{app:protocol}

\paragraph{KDD Cup 1999.} The full \texttt{kddcup.data} file (4{,}898{,}431 records). The three categorical features (protocol, service and flag) are one-hot encoded, which gives 122 dimensions. Removing exact duplicates leaves 1{,}074{,}974 points.

\paragraph{US Census 1990.} 68 coded attributes (the record identifier \texttt{caseid} is dropped); 2{,}458{,}285 records, of which 1{,}998{,}492 are unique.

\paragraph{YearPredictionMSD.} 90 timbre features (the target year is dropped); 515{,}345 records, of which 515{,}129 are unique.

\paragraph{GeoNames.} Latitude and longitude from the \texttt{allCountries} file (13{,}199{,}186 unique coordinate pairs). We draw a random subsample of 1{,}020{,}000 points.

\paragraph{Covertype.} The 54 attributes from scikit-learn's \texttt{fetch\_covtype}: 581{,}012 records, with no duplicates.

\paragraph{Household power.} The 7 numeric columns of the UCI file (date and time are dropped). Rows with missing values are dropped, which leaves 2{,}049{,}280 records, of which 1{,}906{,}698 are unique.

\paragraph{Statlog Shuttle.} The training and test files combined, with the class column dropped: 9 numeric attributes and 58{,}000 records, all unique.

\paragraph{Splits and scaling.} After duplicates are removed, 20{,}000 points are held out (random permutation with seed 1234), and the test set is one half of them (seed 1234). Scaling uses statistics of the training points only.

\paragraph{Embedding datasets.} DEEP~1M uses the e-means paper's split, verified against the SHA-256 checksums distributed with the library's e-means benchmarks. SIFT~1M uses the first $10^6$ BIGANN base vectors for training, and its test set is half of the next $10^4$.

\paragraph{Software and hardware.} FAISS~1.16.0 (\texttt{faiss-cpu} from PyPI), NumPy~2.5.4 and scikit-learn~1.9.1. Two AMD Threadripper servers (3990X and 7980X), each running 8 runs at a time with 4 threads per run.

\paragraph{Timing machine.} The running times of \cref{sec:exp-cost} were measured on the 3990X server: an AMD Ryzen Threadripper 3990X processor (64 cores, 128 hardware threads, 256~MiB of L3 cache), 188~GiB of memory, Ubuntu~24.04.5~LTS with Linux kernel~7.0, Python~3.12.16, NumPy~2.5.4 with OpenBLAS~0.3.34, and FAISS~1.16.0 (\texttt{faiss-cpu} from PyPI). Each timed run used 16 threads at reduced scheduling priority (\texttt{nice}~10). The server is shared with other users, whose jobs we did not control; rotating the order of the three methods across repetitions spreads the effect of their load over the three methods.

\paragraph{e-means settings.} The library's defaults: the scaling laws, \texttt{batch\_law="auto"} (the smaller batch size for $d\le16$) and the last-epoch mean. Every method uses the same settings; only the revival rule differs.

\end{document}